\documentclass[lettersize,journal,twoside]{IEEEtran}
\usepackage{amsmath,amsfonts}
\usepackage[ruled,linesnumbered]{algorithm2e}
\usepackage{array}
\usepackage{multirow}
\usepackage{textcomp}
\usepackage{stfloats}
\usepackage{url}
\usepackage{verbatim}
\usepackage{graphicx}
\usepackage{cite}
\usepackage{color}
\usepackage{subfigure}
\usepackage{bbding}

\usepackage{footnote}

\usepackage{dsfont}
\usepackage{amsfonts,amsmath}
\usepackage{url,cite}
\usepackage{color}
\usepackage{booktabs}
\usepackage{extarrows}
\usepackage[colorlinks,
linkcolor=blue,
anchorcolor=blue,
citecolor=green]{hyperref}

\def\bs{\boldsymbol}

\begin{document}

\title{A Conditional Diffusion Model for Electrical Impedance Tomography Image Reconstruction}

\author{{Shuaikai~Shi, Ruiyuan~Kang and 
Panos~Liatsis,~\IEEEmembership{Senior Member,~IEEE}} \\
	\thanks{This research was supported by ASPIRE UAE under grant AARE20-154.}
	\thanks{Shuaikai Shi, Panos Liatsis are with the Department of Computer Science, Khalifa University, Abu Dhabi, United Arab Emirates (e-mail:shuaikai.shi@ku.ac.ae; panos.liatsis@ku.ac.ae), Ruiyuan Kang is with the Wave and Machine Intelligence Department, Technology Innovation Institute, Abu Dhabi, United Arab Emirates (e-mail: ruiyuan.kang@tii.ae.)}

}

\markboth{Submission to  IEEE Transactions on Instrumentation and Measurement,~Vol.~XX, No.~XX, ~2024}%
{Shi \MakeLowercase{\textit{et al.}}:  A Conditional Diffusion Model for Electrical Impedance Tomography Image Reconstruction }


\maketitle

\begin{abstract}
Electrical impedance tomography (EIT) is a non-invasive imaging technique, capable of reconstructing images of the electrical conductivity of tissues and materials. It is popular in diverse application areas, from medical imaging to industrial process monitoring and tactile sensing, due to its low cost, real-time capabilities and non-ionizing nature. EIT visualizes the conductivity distribution within a body by measuring the boundary voltages, given a current injection. However, EIT image reconstruction is ill-posed due to the mismatch between the under-sampled voltage data and the high-resolution conductivity image. A variety of approaches, both conventional and deep learning-based, have been proposed, capitalizing on the use of spatial regularizers, and the paradigm of image regression. In this research, a novel method based on the conditional  diffusion model for EIT reconstruction is proposed, termed CDEIT. Specifically, CDEIT consists of the forward diffusion process, which first gradually adds Gaussian noise to the clean conductivity images, and a reverse denoising process, which learns to predict the original conductivity image from its noisy version, conditioned on the boundary voltages. Following model training, CDEIT applies the conditional reverse process on test voltage data to generate the desired conductivities. Moreover, we provide the details of a normalization procedure, which demonstrates how EIT image reconstruction models trained on simulated datasets can be applied on real datasets with varying sizes, excitation currents and background conductivities. Experiments conducted on a synthetic dataset and two real datasets demonstrate that the proposed model outperforms state-of-the-art methods. The CDEIT software is available as open-source  (\url{https://github.com/shuaikaishi/CDEIT}) for reproducibility purposes.

\end{abstract}

\begin{IEEEkeywords}
Electrical impedance tomography, image reconstruction, diffusion model, probabilistic model, measurement visualization.
\end{IEEEkeywords}

\section{Introduction}
\label{sec.sec1}
\IEEEPARstart{T}{omographic} imaging is a technique used to reconstruct the internal structure of an object by scanning it from multiple angles\cite{eitreview1}. It is widely applied to the fields such as medicine \cite{appMed}, \cite{appmed1}, \cite{appmed2}, industrial inspection \cite{appInd}, robotics \cite{approb1}, \cite{approb2} and geological exploration \cite{appGeo}. Conventional tomographic imaging methods, such as Computed Tomography (CT) and Magnetic Resonance Imaging (MRI), achieved significant success in terms of accuracy and resolution but also face challenges, e.g., high equipment costs and long imaging times \cite{nextgensensing,detect1}, as well as vulnerability in the visible band \cite{detect2}. 

Electrical Impedance Tomography (EIT), an emerging imaging technology, has garnered extensive attention benefiting from its non-invasive nature, low cost, and real-time imaging capabilities\cite{eitreview2}, and has been applied to fields such as medical imaging, process tomography, nondestructive testing \cite{eitbook}, etc.  EIT infers the impedance distribution inside an object by measuring the voltage changes between electrodes, which could be used for the object structure inversion.
However, limited by the highly nonlinear and ill-posed nature of this inverse problem, conventional EIT image reconstruction methods usually provide suboptimal imaging quality and low resolution\cite{mlmethods}. Recently, Deep Learning (DL)-based reconstruction methods demonstrated the ability to directly map voltage signals to desired conductivity images due to its powerful expressive ability \cite{eitreview3}.

\subsection{Motivation}
Classical image reconstruction methods based on linear back-projection \cite{backprojection} hypothesize a linear model between the conductivity distribution and associated voltage measurements. Additionally, spatial prior knowledge is incorporated by employing regularization \cite{TR} to improve the quality of reconstruction. This type of method can also be applied within an iterative framework for solving the nonlinear inverse problem\cite{GN}. However, in practice, conventional optimization approaches for EIT image reconstruction demonstrated limited performance as they are primarily based on simplified linear modeling, and cannot meet the demands of real-time imaging.

DL-based methods have been employed to address these limitations. These methods use deep neural networks, such as generative adversarial networks (GAN), convolutional neural networks (CNN) and attention modules, to either directly learn the mapping from the voltage signals to conductivity images or indirectly, or improve initial conductivity estimate obtained from conventional methods. The related researches have demonstrated high-definition reconstructions of conductivity images\cite{eitreview2,eitreview3}.
However, the majority of approaches map the voltage information to the output conductivity in a single step. This process can be further improved by incorporating multiple iterative models.
Furthermore, regression-based methods may exhibit limited generalization ability on noisy data due to overfitting on training metrics.
Notably, these issues can be mitigated by integrating generative models \cite{srgan}, such as the recently prevailing diffusion model \cite{ddpm2,ddpm}. 
Motivated by these insights, we propose a novel multi-stage EIT image reconstruction model.

\subsection{Methodology Overview and Contributions}
Recently, the denoising diffusion probabilistic model (DDPM) \cite{ddpm} has garnered significant attention within the deep generative models community and has been applied to various generative tasks, such as image generation \cite{dhariwal2021diffusion}, and audio synthesis \cite{chen2020wavegrad}.
Moreover, DDPM with additional inputs can be employed for conditional generation tasks, including image superresolution \cite{sr3}, image fusion \cite{shi2023hyperspectral}, text-to-image generation \cite{stableDiffusion}, and image editing \cite{imageEdit}, encompassing inpainting, uncropping, etc.

DDPM learns to generate a clean output from its noisy counterpart through a series of denoising steps. Specifically, DDPM incorporates two processes, i.e., the forward diffusion process and the reverse denoising process. During the training phase, the diffusion process introduces independent Gaussian noise to the initial clean image sample multiple times, guiding the output towards a standard Gaussian distribution. Subsequently, the denoising process, implemented by a deep neural network, learns to map noisy input data back to the original clean data. 
After completing training, DDPM can generate new samples by applying the denoising process to Gaussian noise inputs \cite{dit}.
Furthermore, the conditional diffusion model introduces a conditional input into the reverse denoising process, which provides the controller in output generation. Such a design reduce the randomness of generation and a highly correlated one could be acquired. Inspired by this benefit, we propose an EIT image reconstruction model, called CDEIT. It
uses the voltage signal as conditional inputs to shape the generation from the Gaussian noise to spatial details of the desired conductivity image.

The main contributions of this work can be summarized as follows:
\begin{enumerate}
	\item Unlike the majority of state-of-the-art EIT image reconstruction models that perform image regression, we adapt the generative diffusion model to address the EIT inverse problem. The proposed approach models a probability distribution of conductivity images directly conditioned on boundary voltages. This design implicitly learns the prior of conductivity images from the data and can benefit from it when sampling from the conditional distribution.

	\item The proposed CDEIT model produces image reconstruction results through multiple iterative denoising steps, distinguishing it from general-purpose, deep learning-based EIT methods. The multi-stage process enhances the quality of the model output by progressively refining the conductivity images, thereby capturing more spatial details and reducing noise. The backbone is a Transformer-based U-net, with an encoder, which uses multi-scale and windowed attention to extract features, connected to a decoder via residuals for information fusion to predict the desired outputs.
	
	\item We propose a generalization framework, based on voltage and current normalization, supporting EIT image reconstruction models, trained on simulated data, to be tested on real-world data. This method can handle real-world scenarios, where the background conductivity and excitation current values differ from the training data. This offers the ability to adapt image reconstruction models across different conditions without the need for retraining.	
\end{enumerate}

The paper is structured as follows. 
Section \ref{sec.sec2} provides a review of related work, encompassing conventional, deep learning, and diffusion-based image reconstruction approaches. Section \ref{sec.sec3} introduces the mathematical framework of the proposed CDEIT model. In Sec.~\ref{sec.3.gen}, we provide the details of the normalization procedure, which supports the generalization of EIT image reconstruction models, trained with simulated data, on real datasets. In Section \ref{sec.sec4}, we present the image reconstruction experiments, conducted on one simulated and two real-world EIT datasets to demonstrate the effectiveness of our proposed model. We also provide quantitative and qualitative comparison with state-of-the-art methods, while discussing the performance of CDEIT in the presence of noise, and a complexity analysis. Last but not least, Section \ref{sec.sec5} concludes the research and discusses to the further exploration of this work.

\begin{figure}[!t]
	\centering
	\includegraphics[width=8.5cm]{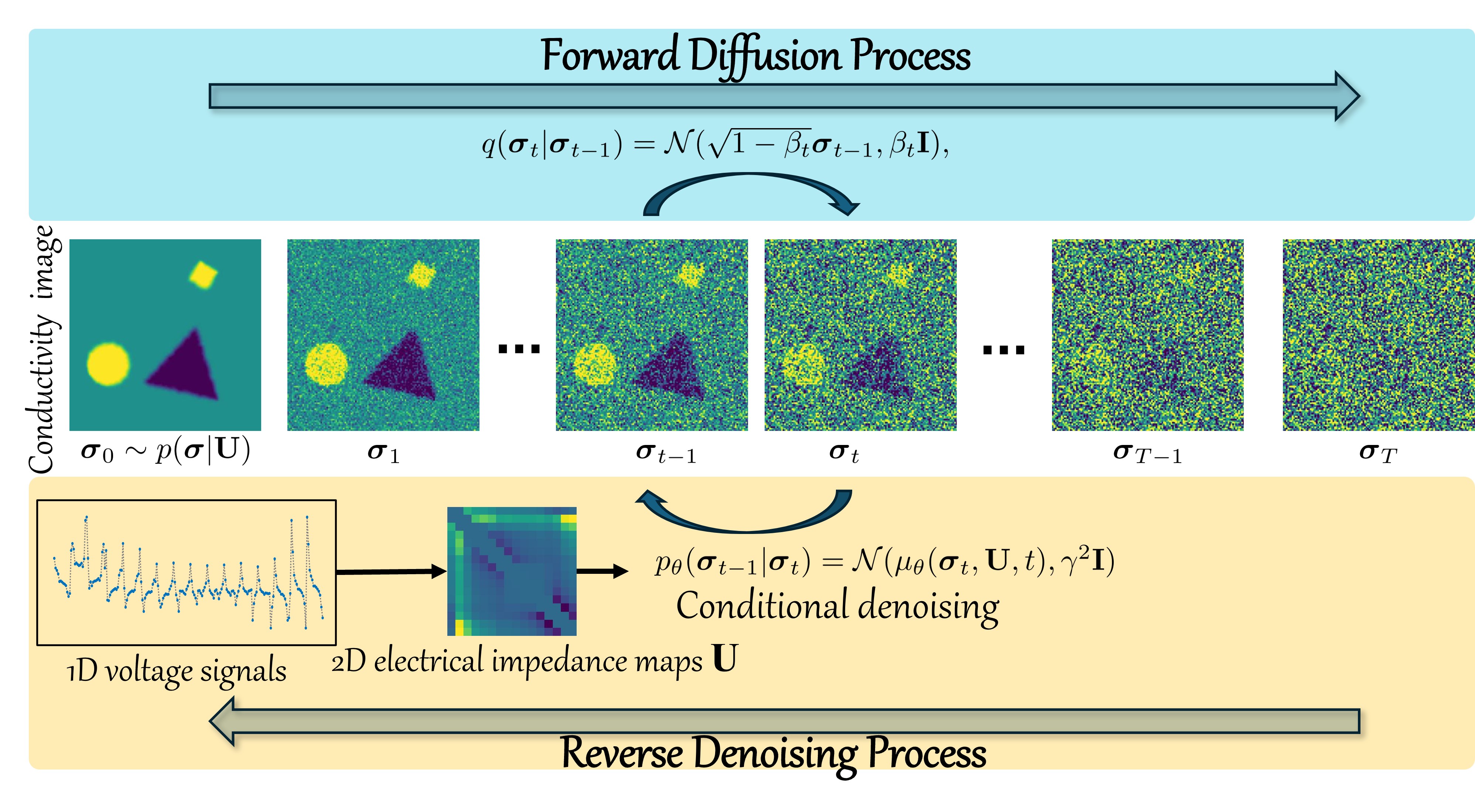}
	\caption{
		Overview of the proposed CDEIT framework, encompassing both the forward and the reverse processes.
		In the forward diffusion process, Gaussian noise is incrementally added to the conductivity image $\bs{\sigma}_0$ over $T$ time steps.
		In the reverse process, the denoising network $\mu_\theta(\cdot)$ progressively restores the spatial details of the original conductivity image, conditioned on the boundary voltages $\mathbf{U}$.
	}
	\label{fig.ddpm}
\end{figure}

\section{Related Work}
\label{sec.sec2}
EIT is considered as an effective tool to obtain the internal conductivity distribution of a target, based on boundary information. In the context of this research, EIT image reconstruction approaches are broadly categorized into conventional methods, which often use a simplified linear model to estimate the conductivity distribution, and  
deep learning approaches, which usually rely on the pairing of conductivity images and corresponding boundary voltage data for supervised learning purposes.

\subsection{Conventional Methods}
Earlier EIT image reconstruction methods were based on the back-projection algorithm \cite{backprojection}. Regularization approaches assist in tackling the ill-posedness. For instance,
Tikhonov regularization (TR) \cite{TR}, also known as ridge regression, assumes that conductivity changes take small values.
In sparse Bayesian learning \cite{sbl}, the $\ell_1$ norm of conductivity changes are introduced to improve the reconstruction accuracy for small targets. The total variation regularizer \cite{tv} is introduced to enforce the reconstruction smoothness. In addition, various physics-informed regularizers have been introduced to improve reconstruction accuracy, such as
Laplacian prior \cite{laplace}, NOSER prior \cite{noser}, etc. These approaches have been used in the context of the single-step and iterative Gauss-Newton (GN) techniques \cite{GN}. Apart from that,
D-bar method\cite{dbar}, a non-iterative reconstruction algorithm, uses a regularization term based on low-pass filtering of the nonlinear backscatter transform of the potential function of the conductivity distribution.

Despite improvements in reconstruction performance, the performance is sensitive to hyperparameter selection, while some approaches require multiple forward model calculations for electric conductivity distribution, resulting in low model efficiency.

\subsection{Deep Learning Approaches}
Deep learning has powerful representation capabilities, enableing it to handle the non-linearity and ill-posed nature of the problem \cite{eitreview2} in a convenient end-to-end manner, and is currently, becoming a new paradigm for EIT reconstruction. 

The improved LeNet \cite{improvedlenet}, which combined CNN and MultiLayer Perceptron (MLP), was first attempted on the EIT reconstruction task. Later on,
Hu \textit{et al.,} introduced a 2D representation of the 1D boundary voltages combined with CNN, namely electrical impedance tomography (CNN-EIM) \cite{cnneim},
which attempts to learn the relationship between spatial variations in conductivity and variations in electrode voltage positions.
Chen \textit{et al.,}  drew on the U-net segmentation network \cite{uneteit}, which is popular in segmentation tasks, to perform EIT image reconstruction, and obtained a significantly better performance, than classical sparse regression models.
A further development was the structure-aware dual-branch network (SADB-Net) \cite{sadbnet}, which incorporates a U-net. In which, one branch of the network is used to segment out the regions of conductivity changes in the target scene, while the other branch extracts the high-level features from the input voltages.The information from the two branches is fused for an accurate estimate of conductivity distribution changes.
Following the work, Yu \textit{et al.,} proposed the structure-aware hybrid-fusion learning (SA-HFL)  model \cite{sahfl} to improve the fusion efficiency of the features from the two branches, which enhances the correlation between the image segmentation results and the predicted conductivity distribution changes.
In addition, the radial basis function was added to CNN (CNN-RBF) network \cite{cnnrbf}, to accelerate network convergence, as well as enhance reconstruction accuracy and robustness. Residual learning is also introduced to the field \cite{ecnet} to improve the initial solutions from conventional methods, and also gain better performance than classical CNNs. Similar idea was also applied to dual-branch U-net (DHU-Net) model \cite{dhunet}. Meanwhile, the deep image prior was introduced to EIT \cite{dipeit} to allow the network to learn a spatial prior for the conductivity distribution to be reconstructed. However, this approach requires multiple finite element calculations for the forward process. 

In summary, DL approaches for the solution of the EIT inverse problem are becoming increasingly complex, leading to a substantial growth in the number of model parameters. Consequently, model training is becoming more computationally expensive, while necessitating the use of substantial amounts of training data to mitigate overfitting.
  
\subsection{Diffusion Models}
The generative diffusion model, inspired by non-equilibrium statistical physics, was first introduced in \cite{ddpm2}. 
Subsequent studies \cite{ddpm} enhanced this model for image generation, leading to the development of the denoising diffusion probabilistic model (DDPM).
An intriguing aspect of this approach is the ability to attach additional conditions to the input data for performing conditional generation tasks. For example, super-resolution tasks can be conditioned on low-resolution images \cite{sr3}, or controlled semantic generation can be conditioned on text \cite{stableDiffusion}. The diffusion process can be applied at the pixel level \cite{ddpm, sr3}, where the dimensions of the latent space are equal to the input data, or at lower dimensions to reduce computational burden and the number of diffusion steps \cite{stableDiffusion}. Due to the multi-stage generation process, DDPM-based models achieved state-of-the-art performance in fields such as image generation and restoration \cite{dhariwal2021diffusion}.
DDPM was also introduced to EIT image reconstruction. 
Wang \textit{et al.,} employed the  score-based diffusion
model (CSD) \cite{csd} to learn the unconditional distribution of high-resolution conductivity images. The image reconstruction estimates from the iterative GN method were then introduced as the sampling midpoint, which was subsequently refined using a noise reduction network. Although this method is capable of obtaining high-resolution conductivity images, it is unable to accurately represent the curvature and sharpness of the boundary at the location of the conductivity change since the learning process of the model is independent on the boundary voltages.
Another approach proposed a diffusion model \cite{ktcdiffusion} combined with the initial guess from several conventional methods. 
Similarly, this model is trained without the use of boundary voltages, resulting in a model performance that relies on the accuracy of the classical approaches.
Direct reconstruction of conductivity images from voltage signals mitigates the cumulative error inherent in classical methods. However, the direct approach presents significant challenges due to the highly ill-posed nature of the inverse problem.

\section{CDEIT: Conditional Diffusion Model-based EIT reconstruction}
In this section, we introduce the proposed CDEIT, covering the problem formulation, forward and backward processes, loss function, optimization techniques, and fast reconstruction strategy.
\label{sec.sec3}

\subsection{Problem Formulation}
The forward problem of EIT is the calculation of the  electric potentials  induced by the   conductivity distribution in the target region. In practice, we are concerned with the relationship between the conductivity values and the associated voltages across the boundary electrodes.
The relationship between these physical quantities in the 2D EIT formulation, when current is injected through a pair of electrodes is given by the following set of equations \cite{eitreview2}:
\begin{figure}[!t]
	\centering
udegraphics[width=8cm]{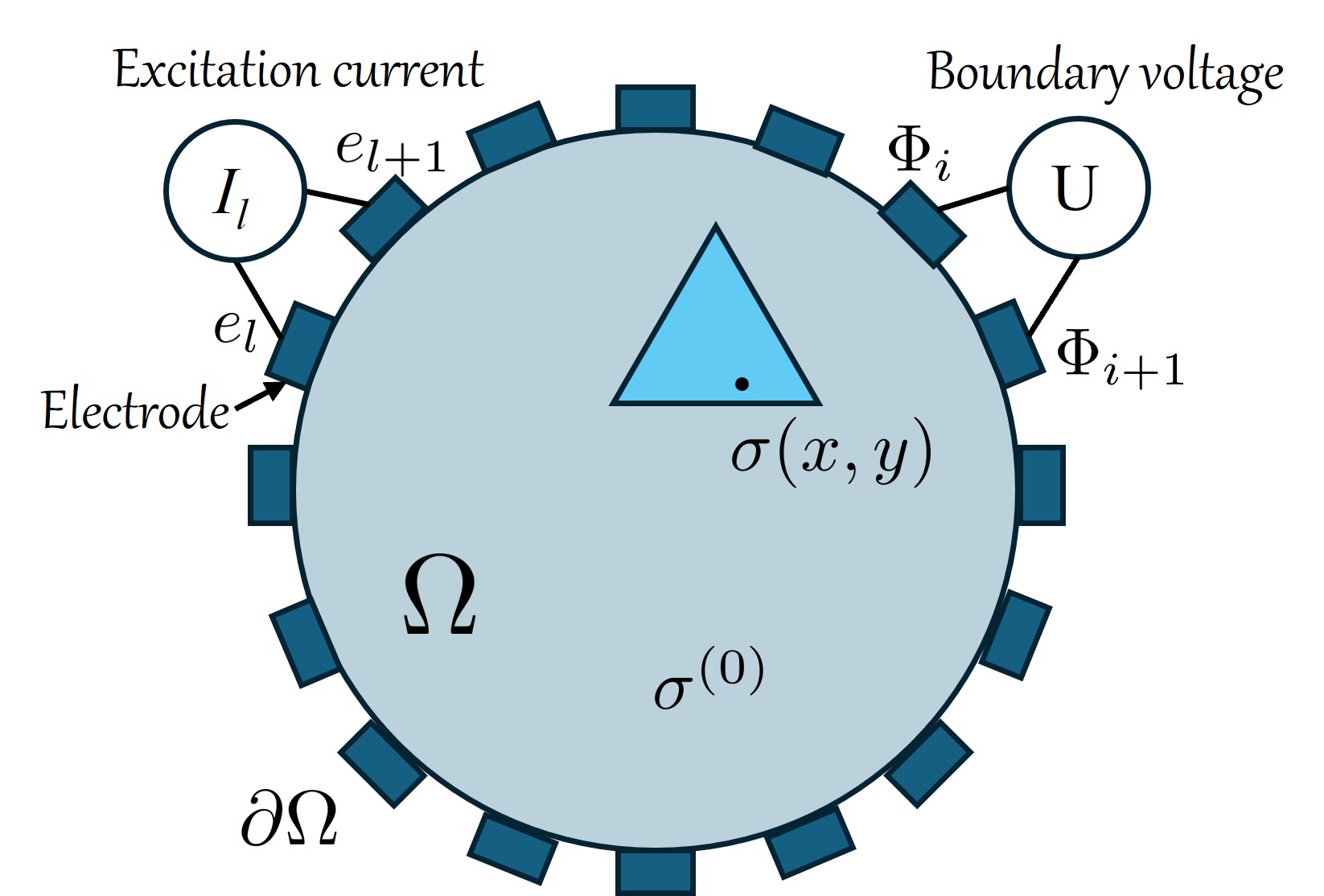}
	\caption{
		2D EIT forward model with 16 electrodes. 
		When a current excitation is applied through a pair of electrodes, changes in the conductivity distribution of the target will result to changes in the voltages of the  boundary electrodes.
	}
	\label{fig.2deit}
\end{figure}
\begin{figure*}[!t]
	\centering
	\includegraphics[width=16cm]{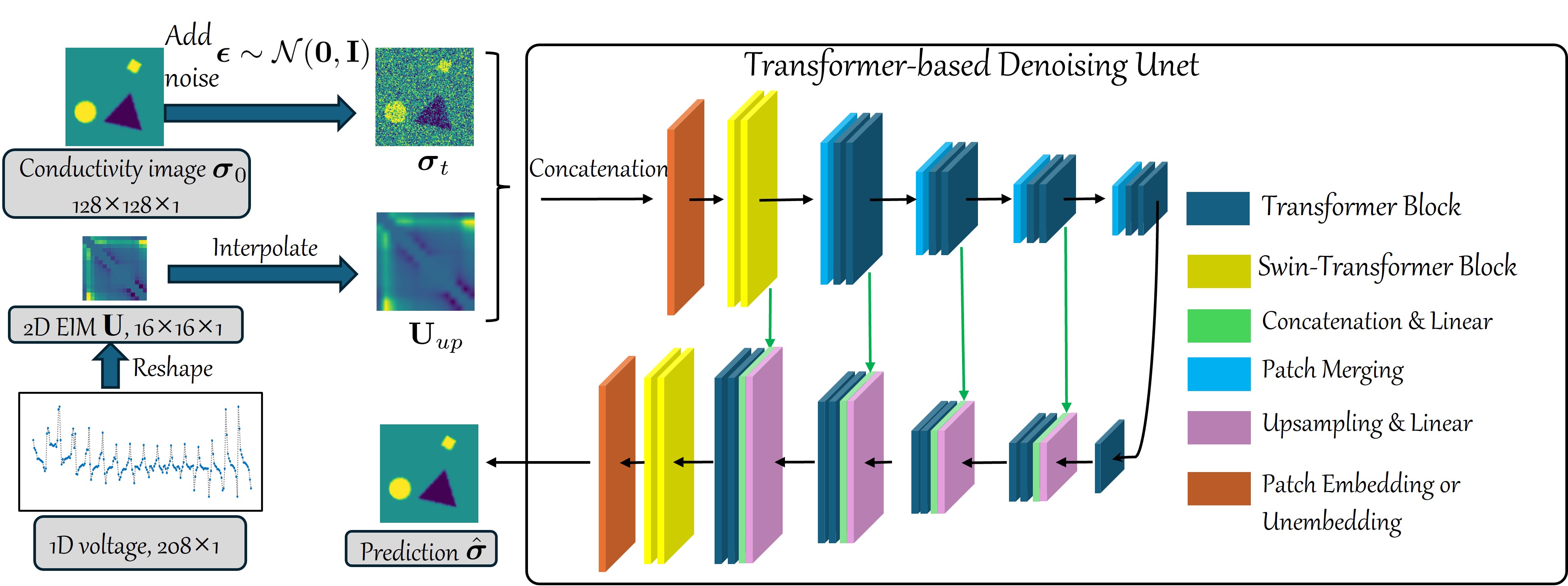}
	\caption{
		Framework of the denoising net $\boldsymbol{\sigma}_\theta(\bs{\sigma}_t,\mathbf{U},t)$.
	}
	\label{fig.ddpm1}
\end{figure*}

\begin{align}
\nabla \cdot \left( {\sigma} \nabla  {\phi}    \right) =0  & \quad in\quad \Omega, \label{eq.eit.f1}\\
 {\phi}  + z_l  {\sigma} \frac{\partial  {\phi}   }{\partial  \bs{n} }=\Phi _l &  \quad on \quad\partial\Omega \\
 \int_{e_l}  \sigma \frac{\partial  {\phi}   }{\partial  \bs{n} }dS =I_l & \quad on \quad\partial\Omega,\\
 \sigma \frac{\partial  {\phi}   }{\partial  \bs{n} }\Big|_{\partial \Omega \backslash\cup e_l} =0  &\quad on \quad\partial\Omega,\label{eq.eit.f4}
\end{align}
where $\sigma,\phi,\bs n$, respectively, denote the conductivity, electric potential and unity outward-pointing
vector at position $(x,y)$ within the area $ \Omega$, $e_l$ is the position of the $l$-th electrode on the boundary $\partial \Omega$, $l=1,2,\dots,L$ is the index of the number of  boundary electrodes, $\Phi_l$ is the  potential on the $l$-th electrode and $I_l$ is the current excitation from the $l$-th electrode. The solution of the forward problem is commonly found using the finite element method (FEM).
Fig. \ref{fig.2deit} depicts a 16-electrode 2D EIT model, which uses the adjacent excitation and adjacent measurement protocols \cite{appmed2}. 
The noise-free forward problem can be represented as:

\begin{equation}
\mathbf{U}=f(\bs\sigma),
\end{equation}
where $\mathbf{U}$ denotes the vector of boundary voltages, $f(\cdot)$ represents the nonlinear forward operator and $\bs\sigma=\{\sigma (x,y)| (x,y)\in \Omega\}$ is the conductivity distribution.

EIT image reconstruction, i.e., an inverse problem,  includes recovering the position and shape of the objects inside the conductivity distribution. This is achieved by solving the following nonlinear least squares problem \cite{approb1}:

\begin{equation}
	\hat{\bs{\sigma}}=\arg \min \limits_{\bs\sigma}\| f(\bs\sigma)-\mathbf{U}\|_2^2.
\end{equation}
 In order to suppress most of the systematic errors and to tolerate small errors in the model, the time-difference EIT model is commonly adopted: 
 
\begin{align}
	\mathbf{U}^{(0)}&=f(\bs\sigma^{(0)})\\
	\mathbf{U}^{(i)}&=f(\bs\sigma^{(i)})
\end{align}

where the estimate of the conductivity distribution changes, $\delta \bs\sigma=\bs\sigma^{(i)}-\bs\sigma^{(0)}$, is obtained from the values of the voltage changes, $\delta \mathbf{U}=\mathbf{U}^{(i)}-\mathbf{U}^{(0)}$ between two sets of measurements.
Typically, $\bs\sigma^{(0)}$ and $\mathbf{U}^{(0)}$ are the conductivity distribution and associated voltage measurements of the background, respectively. Similarly, $\bs\sigma^{(i)}$ and $\mathbf{U}^{(i)}$ are the conductivity distribution and associated voltage measurements at instance $i$, when the inclusions are present, respectively. 
In the text that follows, unless otherwise specified, all voltage and conductivity quantities will relate to time differences.

In summary, the inverse EIT problem aims to recover the conductivity distribution $\bs\sigma$ using the information of the voltage measurement $\mathbf{U}$. In the context of this research, EIT image reconstruction is posed as the problem of modeling the conditional probabilistic distribution $p(\bs\sigma|\mathbf{U})$. In this perspective, it is possible to obtain the desired conductivity distribution through sampling this conditional distribution.
Drawing inspiration from the advanced denoising diffusion generative model \cite{ddpm}, in the following subsections, we describe the specifics of the two processes to learn the aforementioned conditional distribution.

\subsection{Forward Diffusion Process}

Building on DDPM \cite{ddpm}, a Markovian chain of length $T$ with a Gaussian diffusion kernel is employed to incrementally add noise to the conductivity values  $\bs\sigma$:
 
\begin{equation}
q(\boldsymbol{\sigma}_t|\boldsymbol{\sigma}_{t-1})=\mathcal{N}(\sqrt{1-\beta_{t}}\boldsymbol{\sigma}_{t-1},\beta _t\mathbf{I}),
\label{eq.diffusionKernel}
\end{equation}
where $t\in \{1,2,\dots,T\}$, $\bs{\sigma}_0=\bs{\sigma}$ is the conductivity ground truth,
$\bs{\sigma}_T$ is the noisy conductivity image after the last forward step,
 $\beta_t \in \{\beta_1,\beta_2,\dots,\beta_T\}$ is a sequence of hyperparameters representing the variances of Gaussian noise and $\mathbf I$ denotes the identity matrix. 

Given the characteristics of the Gaussian distribution, it is possible to directly determine the output at each diffusion step by:
\begin{align}
	q(\bs{\sigma}_t|\bs{\sigma}_{0})&=\mathcal{N}(\sqrt{\bar\alpha_t}\bs{\sigma}_{0},\sqrt{1-\bar\alpha_t}\mathbf{I}) 	\label{eq.forwardGaussian}\\ 
	\text{or}\; \bs{\sigma}_t&=\sqrt{\bar {\alpha}_t}\bs{\sigma}_0 + \sqrt{1-\bar {\alpha}_t} \boldsymbol{\epsilon},
	\label{eq.forwardGaussianC}
\end{align}
where $\bar {\alpha}_t=\prod_{s=1}^{t}(1-\beta_s) $ and $\boldsymbol{\epsilon}\sim \mathcal{N}(\mathbf{0},\mathbf{I})$.
Note that when $T$ approaches infinity, 
$q(\bs{\sigma}_ \infty)$ converges to $\mathcal{N}(\mathbf{0},\mathbf{I})$. The forward diffusion process is illustrated at the top of Fig.~\ref{fig.ddpm}.

\subsection{Reverse Denoising Process}

Following the completion of the forward diffusion process, the reverse denoising process is applied to reconstruct the desired conductivity image from the noise input and the information of the corresponding boundary voltages, as shown at the bottom of Fig.~\ref{fig.ddpm}.

In particular, the reverse process transforms the standard Gaussian noise into the clean conductivity image $\bs{\sigma}$. It is assumed that the inverse of the previously described forward diffusion step \eqref{eq.diffusionKernel} follows another Gaussian kernel, as detailed below:
 
\begin{equation}
	p_\theta(\bs{\sigma}_{t-1}|\bs{\sigma}_t)=\mathcal{N}(\mu_\theta(\bs{\sigma}_t,\mathbf{U},t),\gamma^2_t\mathbf{I}),
	\label{eq.inverseKernel}
\end{equation}
where $\mu_\theta(\cdot)$ is a function modeled by a deep neural network parameterized by $\theta$ and
 $\gamma^2_t $ represents  a sequence of hyperparameters $ \{\gamma^2_1,\gamma^2_2,\dots,\gamma^2_T\}$. The specific values for these hyperparameters will be discussed in the next subsection.

\subsection{Loss Function}
\label{sec.subsec3objFunc}
In this subsection, we introduce the objective function of the proposed CDEIT model. The noisy conductivity image from the forward process is considered as a sequence of hidden variables and a variational inference framework is employed to derive the loss function.
The evidence lower bound (ELBO) of the log-likelihood is given by \cite{bound}:
\begin{equation}
	\mathcal{L}(\theta)=-\sum_{t=1}^T \text{KL}\left(q(\bs{\sigma}_{t-1}|\bs{\sigma}_{t},\bs{\sigma}_{0})||p_\theta(\bs{\sigma}_{t-1}|\bs{\sigma}_{t})\right),
	\label{eq.elboKL}
\end{equation}
where $\text{KL}(\cdot||\cdot)$ is the Kullback-Leibler divergence between the conditional posterior distribution and \eqref{eq.inverseKernel}.
A comprehensive derivation of \eqref{eq.elboKL} can be found in the Appendix.
$q(\bs{\sigma}_{t-1}|\bs{\sigma}_{t},\bs{\sigma}_{0})$ is the conditional posterior distribution which can be derived as:

\begin{align}
q(\bs{\sigma}_{t-1}|\bs{\sigma}_{t},\bs{\sigma}_{0})&\propto q(\bs{\sigma}_{t}|\bs{\sigma}_{t-1},\bs{\sigma}_{0})q(\bs{\sigma}_{t-1}|\bs{\sigma}_{0})\nonumber\\
&=q(\bs{\sigma}_{t}|\bs{\sigma}_{t-1})q(\bs{\sigma}_{t-1}|\bs{\sigma}_{0}).
\label{eq.conditionalPosterior}
\end{align}
Since both the conditional distribution \eqref{eq.diffusionKernel} and the prior distribution \eqref{eq.forwardGaussian} are Gaussian, the posterior distribution inherits this Gaussian form. This is due to the self-conjugate property \cite{prml} of the Gaussian distribution.  
Thus, the distribution \eqref{eq.conditionalPosterior} described above is:
\begin{equation}
q(\bs{\sigma}_{t-1}|\bs{\sigma}_{t},\bs{\sigma}_{0}) =\mathcal N (\tilde\mu_t(\bs{\sigma}_t,\bs{\sigma}_0),
\tilde \beta_t \mathbf{I}),
\label{eq.reverse}
\end{equation}
where 
\begin{align}
	\tilde\mu_t(\bs{\sigma}_t,\bs{\sigma}_0)&= \frac{\sqrt{1-\beta_t}(1-\bar \alpha _{t-1})}{1-\bar \alpha _t}\bs{\sigma}_t+\frac{\sqrt {\bar \alpha_{t-1}}\beta_t}{{1-\bar \alpha _t}} \bs{\sigma}_0,\label{eq.predMean}\\
	\tilde \beta_t&=\frac{1-\bar\alpha_{t-1}}{1-\bar\alpha_{t}}\beta_t.
\end{align}

The KL divergence in \eqref{eq.elboKL} is tractable since it only involves the diagonal Gaussian distributions. Consequently, the KL divergence for a single time step can be expressed as:
\begin{align}
	\mathcal {L}_t(\theta)
	&=-\text{KL}(q(\bs{\sigma}_{t-1}|\bs{\sigma}_{t},\bs{\sigma}_{0})||p_\theta(\bs{\sigma}_{t-1}|\bs{\sigma}_{t},\mathbf{U}))\nonumber\\
	&=-\frac{1}{2\gamma^2_t}\|\tilde\mu_t (\bs{\sigma}_t,\bs{\sigma}_0)-\mu_\theta(\bs{\sigma}_t,\mathbf{U},t)\|^2_2+C,
	\label{eq.klOneStep}
\end{align}
where $C$ is a constant independent of $\theta$ and includes certain fixed hyperparameters.

By further simplification, the network prediction  $\mu_\theta(\bs{\sigma}_t,\mathbf{U},t)$  can be written in a form similar to  \eqref{eq.predMean}, i.e.,: 
\begin{align}
	&\mu_\theta(\bs{\sigma}_t,\mathbf{U},t)\nonumber\\
= 	&\frac{\sqrt{1-\beta_t}(1-\bar \alpha _{t-1})}{1-\bar \alpha _t}\bs{\sigma}_t+\frac{\sqrt {\bar \alpha_{t-1}}\beta_t}{{1-\bar \alpha _t}} \bs{\sigma}_\theta(\bs{\sigma}_t,\mathbf{U},t),
	\label{eq.reverseFusion}
\end{align}
where $\boldsymbol{\sigma}_\theta(\cdot)$ is an alternative expression of $\mu_\theta(\cdot)$, which estimates the clean conductivity image during the forward process and will be implemented in the following subsection.

Consequently, the Kullback-Leibler divergence presented in  \eqref{eq.klOneStep} can be further streamlined:
\begin{equation}
	\mathcal {L}_t(\theta)=- \frac{\bar\alpha_{t-1}\beta_t }{2  (1-\bar\alpha_t)^2}
	\|\boldsymbol{\sigma}_0-\bs{\sigma}_\theta(\bs{\sigma}_t,\mathbf{U},t)\|^2_2
\end{equation}
During the training phase, we establish the relationship $\gamma^2_t=\tilde \beta_t$. Subsequently, we can optimize the  ELBO  as expressed in \eqref{eq.elboKL} through a stepwise process along the time dimension. At each step, we concentrate on a simplified loss function, disregarding the hyperparameter coefficients, which can be formulated as: 
\begin{equation}
	\mathcal {L}_{\text{simple}}(\theta)=  
	\|\boldsymbol{\sigma}_0-\bs{\sigma}_\theta(\bs{\sigma}_t,\mathbf{U},t)\|_1
	\label{eq.lossSimple}
\end{equation}
For our experimental protocol, 
the $\ell_1$ loss function is used to replace the $\ell_2$ loss function due to its improved robustness to outliers.

\subsection{Network \& Optimization}
Next,  the conditional denoising network $\boldsymbol{\sigma}_\theta(\cdot)$ is constructed using the Transformer-based U-net architecture. 
Overall, the network is shown in Fig.~\ref{fig.ddpm1}.
The inputs to the U-net consists of three parts, $\{\bs{\sigma}_t,\mathbf{U},t\}$.
At each training step $t$, Gaussian noise is added to the clean conductivity image $\bs{\sigma}_0$ to produce  $\bs{\sigma}_t$ via \eqref{eq.forwardGaussianC} in the forward diffusion process.
By listing the results of the voltages produced by each current excitation in one row and by setting the voltage values associated with the excitation electrodes to zero, a 2D electrical impedance maps (EIM) $\mathbf{U}$ can be constructed that contains some spatial information.
Interpolation is applied to the EIM  to match the spatial resolution of the input image.
Subsequently, $\bs{\sigma}_t$
and the upsampled 
$\mathbf{U}_{up}$
are concatenated along the channel dimension.
These data preprocessing steps can be represented as
\begin{align}
	\mathbf{U}_{up}&=\text{Interp.} (\mathbf{U}) \uparrow_{S\times},\\
	\mathbf{X}&=[\bs{\sigma}_t,\mathbf{U}_{up}].
\end{align}
In the experiments, the spatial resolution of the conductivity image $\bs{\sigma}$ is $128\times 128$, which is divisible by the number of electrodes, i.e., $16$.
The symbol $\uparrow_{S\times}$ denotes an increase in the spatial resolution of the 2D EIM by a factor of 
$S$ through nearest neighbor interpolation. 
The notation $[\cdot]$ signifies the concatenation operation, while $\mathbf{X}$ represents the data input fed into the denoising network.
The input is embedded as a sequence with a patch size of 2 and a model dimension of 512, then processed by subsequent transformer-based blocks.

\begin{figure}[!t]
	\centering
	\subfigure[  ]{
		\includegraphics[width=4cm]{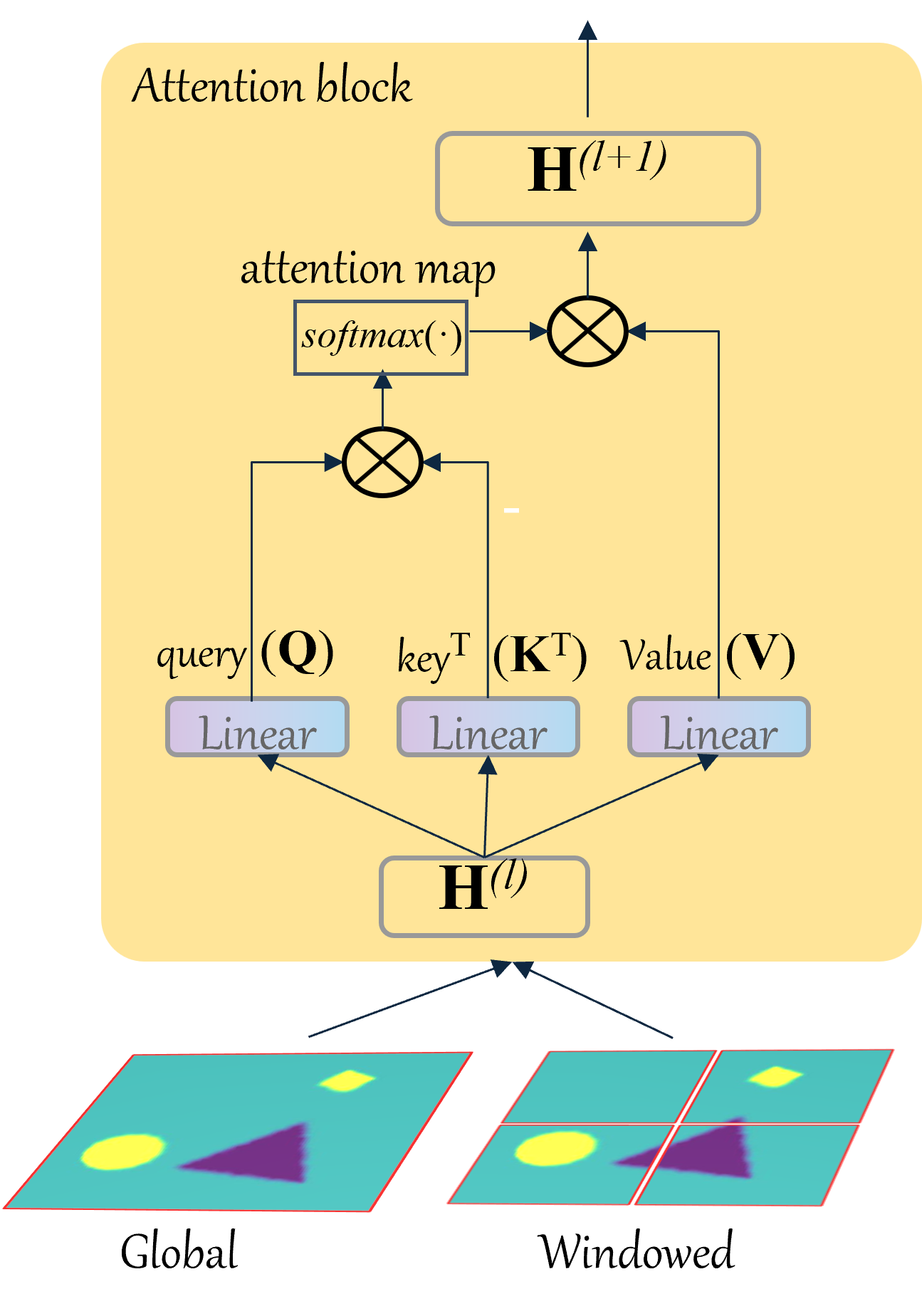}}
	\subfigure[  ]{
		\includegraphics[width=4cm]{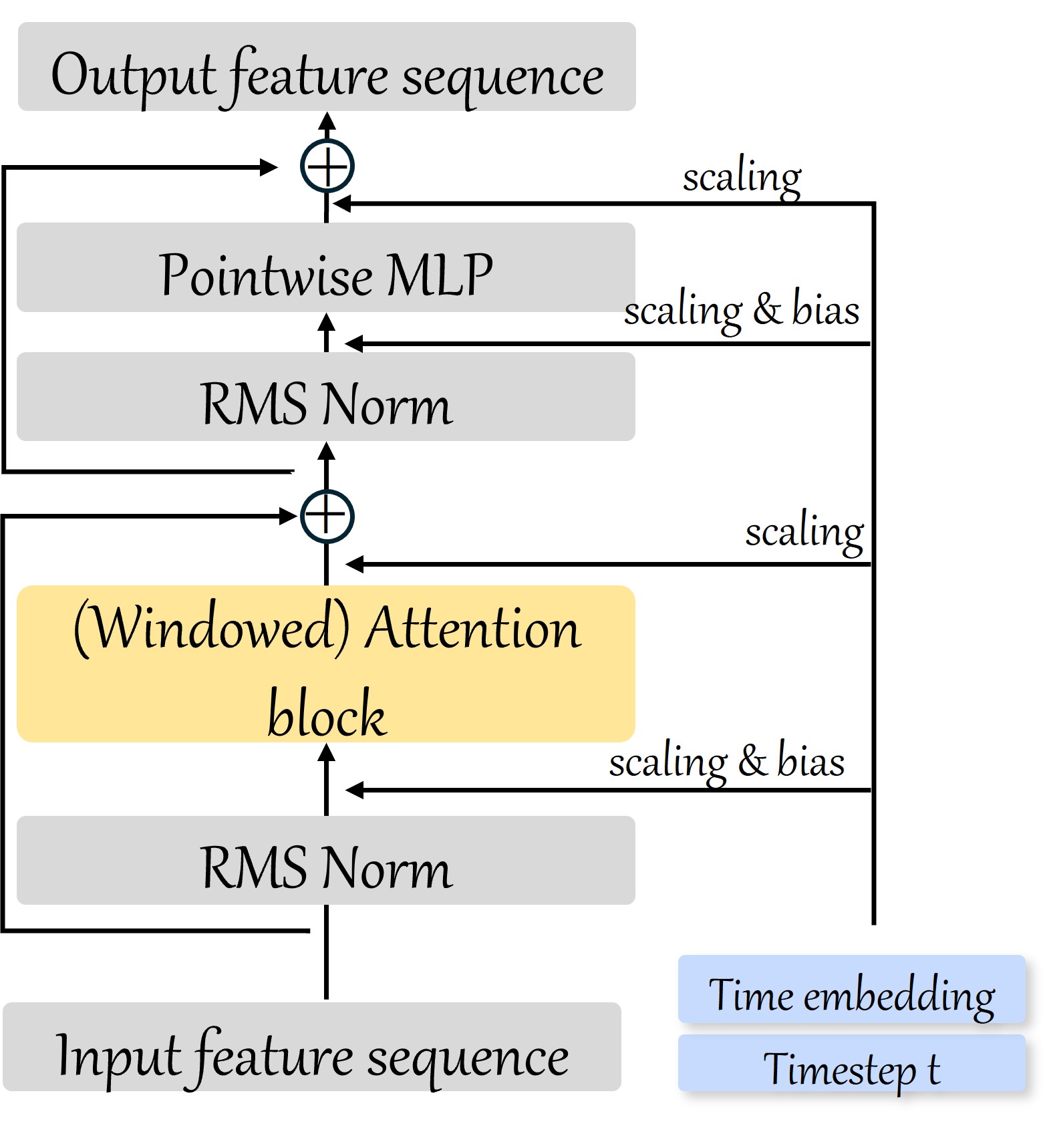}}
	\caption{
		Transformer block and Swin-Transformer block. (a) global attention and windowed attention. (b) Time-involved Transformer blocks.
	}
	\label{fig.transformer2}
\end{figure}

In addition, the Transformer-based denoising U-net comprises a series of Transformer blocks, Swin-Transformer blocks, patch merging, skip connections, and upsampling modules. 
The Transformer block computes attention globally on the feature map, while the Swin-Transformer splits the map into small patches to compute attention inside the window in a memory-saving manner, as shown in Fig.~\ref{fig.transformer2} (a). After feature embedding, the result of $l$-th layerself-attention mechanism, $\mathbf{H}^{(l+1)}$, is computed as
\begin{equation}
    \mathbf{H}^{(l+1)}=\text{Softmax}(\frac{\mathbf{QK}^\mathrm{T}}{\sqrt{d_k}})\mathbf{V} ,
\end{equation}
where $d_k$ is the dimension of the key as a normalization factor, and $\mathbf{Q}$, $\mathbf{K}$ and $\mathbf{V}$  represent the query, key and value, respectively.
These are computed by the linear projections of the input features $\mathbf{H}^{(l)}$ with the model parameters $\mathbf{W}_q,\mathbf{W}_k$ and $\mathbf{W}_v$, i.e., $\mathbf{Q}=\mathbf{H}^{(l)}\mathbf{W}_q $,
$\mathbf{K}=\mathbf{H}^{(l)}\mathbf{W}_k $ and 
$\mathbf{V}=\mathbf{H}^{(l)}\mathbf{W}_v $.
The current time step $t$ is encoded as a fully learned  time embedding  as the positional embedding in the Transformer \cite{transformer} and inputs to each Transformer and Swin-Transformer block \cite{liu2021Swin} within the network as affine transform coefficients as shown in Fig.~\ref{fig.transformer2} (b).

\begin{figure}[!t]
	\centering
	\subfigure[  ]{
		\includegraphics[width=4cm]{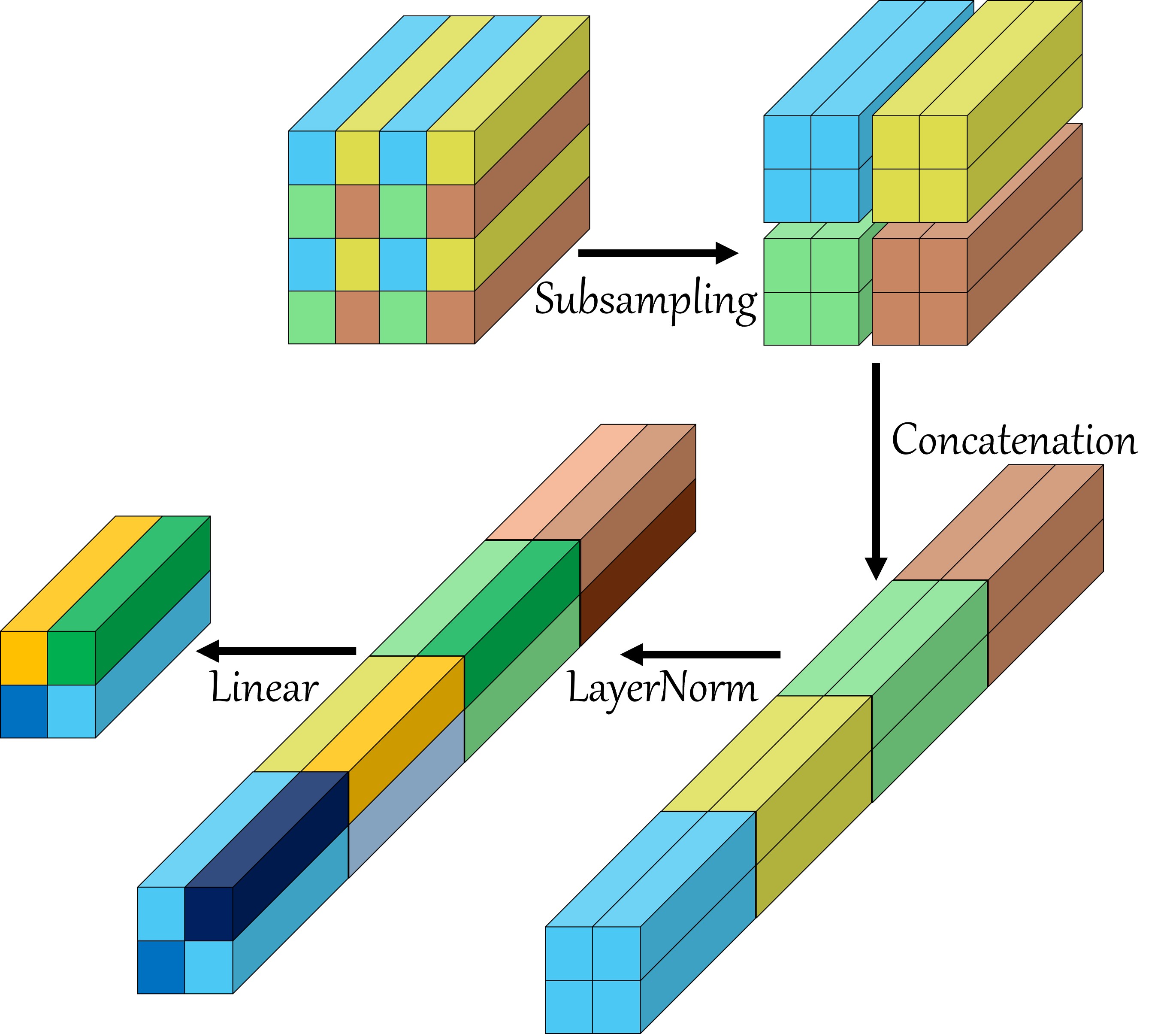}}
	\subfigure[  ]{
		\includegraphics[width=4cm]{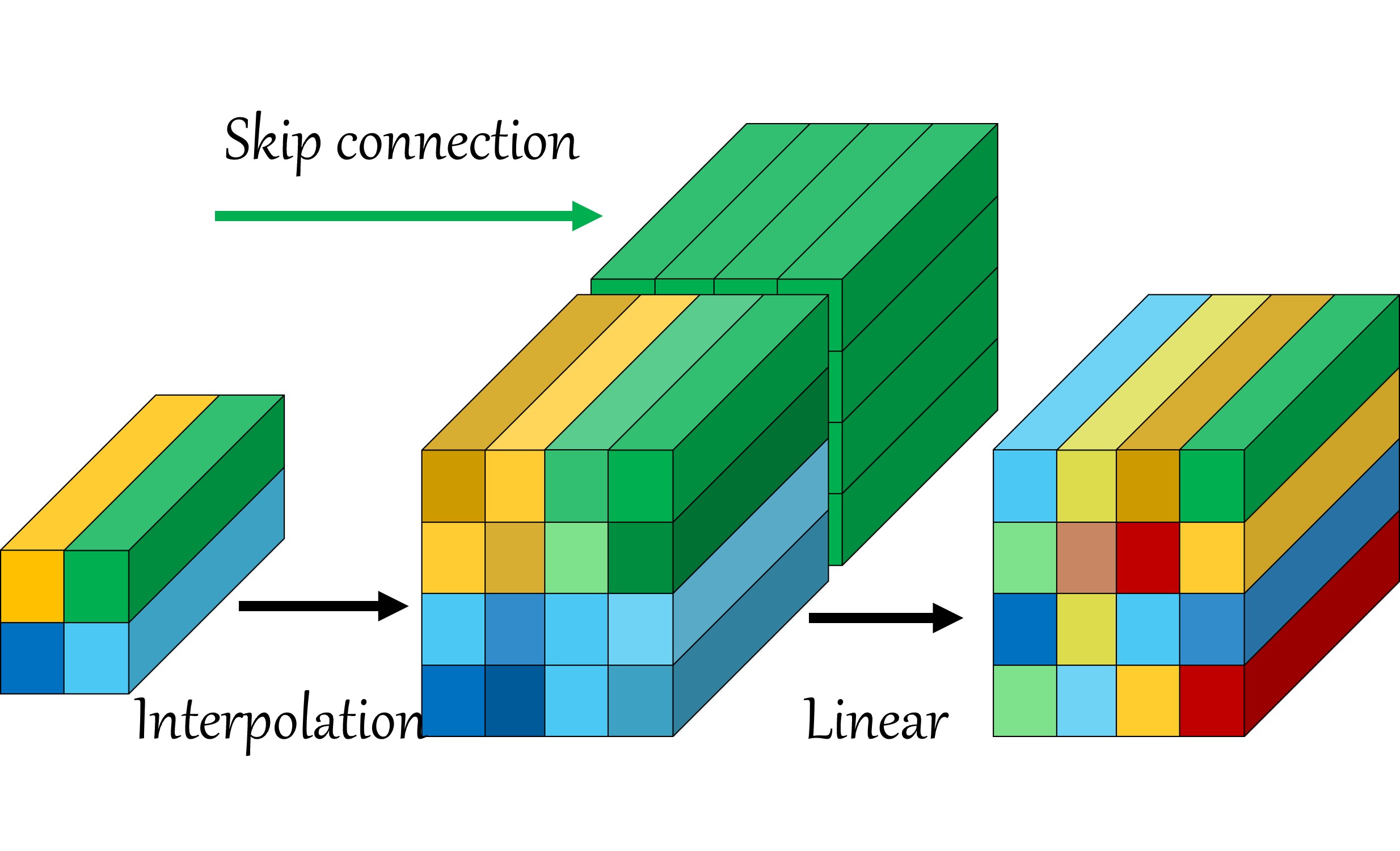}}
	\caption{
		Scale transformation modules. (a) Patch merging. (b) Upsampling by interpolation.
	}
	\label{fig.patchmerge}
\end{figure}

The patch merging is employed to downsample the feature map in the upper encoder, preserving information that might otherwise be lost through conventional pooling operations, while the upsampling module comprises the nearest neighbor  interpolation and a linear layer for feature channel adjustment as shown in Fig.~\ref{fig.patchmerge}.
Finally, the network outputs an estimate of the clean conductivity image. To adapt the model for EIT image reconstruction, we employ the Sigmoid-weighted Linear Unit (SiLU) nonlinear activation function \cite{swish} and root mean square (RMS) normalization \cite{rmsnorm}. The training process of the proposed CDEIT is outlined in Algorithm \ref{al.DDPM}.

\begin{algorithm}[!t]
	\small
	\SetAlgoLined
	\caption{Training algorithm of CDEIT}
	\label{al.DDPM}
	\LinesNumbered  
	\KwIn{paired training data: ($\mathbf{U}$, $\bs{\sigma}_0$)\;
		hyperparameter sequence: $\{\beta_1,\beta_2,\dots,\beta_T\}$\; 
        model parameters: $\theta$.
	}
	Initialize $\theta$ randomly\;
	\Repeat{\textbf{training phase end}}{
		$\boldsymbol{\epsilon}\sim \mathcal{N}(\mathbf{0},\mathbf{I})$\;
		$t\sim U(1,T)$\;
		Compute the noisy image through \eqref{eq.forwardGaussianC}: $\bs{\sigma}_t=\sqrt{\bar {\alpha}_t}\bs{\sigma}_0 + \sqrt{1-\bar {\alpha}_t} \boldsymbol{\epsilon}$\\
		Compute the gradient of \eqref{eq.lossSimple} w.r.t. $\theta$:
		$\nabla_\theta\|\boldsymbol{\sigma}-\boldsymbol{\sigma}_\theta(\bs{\sigma}_t,\mathbf{U},t)\|$\;
		Update ${\theta}$ via Adam optimizer \cite{adam}.\\
	}
	
\end{algorithm}

\begin{algorithm}[!t]
	\small
	\SetAlgoLined
	\caption{Reconstruction of CDEIT }
	\label{al.DDPMtest}
	\LinesNumbered  
	\KwIn{test boundary voltage: $\mathbf{U}^\prime$\;
		the sub-sequence of $\{1,2,\dots,T\}$ is denoted as $\{\tau _1,\tau_2,\dots,\tau_d =T\}$
	}
	
	$\bs{\sigma}_T^\prime\sim \mathcal N(\mathbf{0},\mathbf{I})$\;
	\For{$t=\tau_d ,\dots,\tau _1$}	 {$\boldsymbol{\epsilon}\sim \mathcal{N}(\mathbf{0},\mathbf{I})$ if $t>1$, else $\boldsymbol{\epsilon}=\mathbf{0}$\\
		Compute the denoised result by $\boldsymbol{\sigma}_\theta(\bs{\sigma}_t^\prime,\mathbf{U}^\prime,t)$\;
		Sample the denoising output $\bs{\sigma}_{t-1}^\prime$ from   DDIM \cite{ddim} sampler \eqref{eq.ddim}\;
	}
\end{algorithm}

\subsection{CDEIT with Fast Sampling}

Upon completion of training, we can generate image reconstruction results on the test data by implementing the reverse denoising process. Rather than sampling from \eqref{eq.inverseKernel} in a step-by-step manner, we employ an alternative reverse sampler to expedite the reconstruction process. This sampler, introduced in \cite{ddim}, is based on a non-Markovian forward process that allows for step-skipping during the reverse process implementation, known as denoising diffusion implicit models (DDIM).
Specifically, the new forward process can be formulated as:

\begin{align}
&q_\sigma (\bs{\sigma}_{t-1}|\bs{\sigma}_{t},\bs{\sigma}_{0})\nonumber\\
=&\mathcal{N}(
\sqrt{\bar{\alpha}_{t-1}}\bs{\sigma}_{0}+ \sqrt{1-\bar\alpha_{t-1}-\gamma_t^2}  \cdot  \frac{\bs{\sigma}_t-\sqrt{\bar{\alpha}}_t\bs{\sigma}_{0}}{\sqrt{1-\bar\alpha_t}},\gamma^2_t \mathbf{I}).
\label{eq.ddim}
\end{align}

Leveraging this relationship, we can utilize the estimated values of $\bs{\sigma}_0$ to execute the reverse process.
Specifically, the fused results on the test data can be generated by implementing the reverse denoising process, as outlined in Algorithm \ref{al.DDPMtest}.

\begin{figure}[!tb]
	\centering

	\subfigure[  ]{
		\includegraphics[height=2.6cm ]{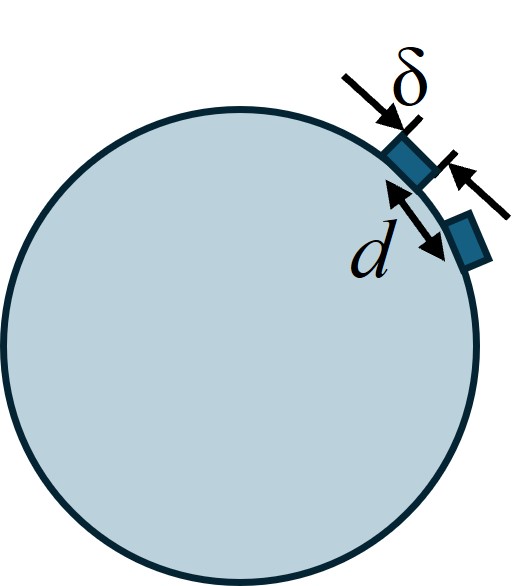}
	}\hspace{1.6cm}
	\subfigure[  ]{
		\includegraphics[height=2.6cm ]{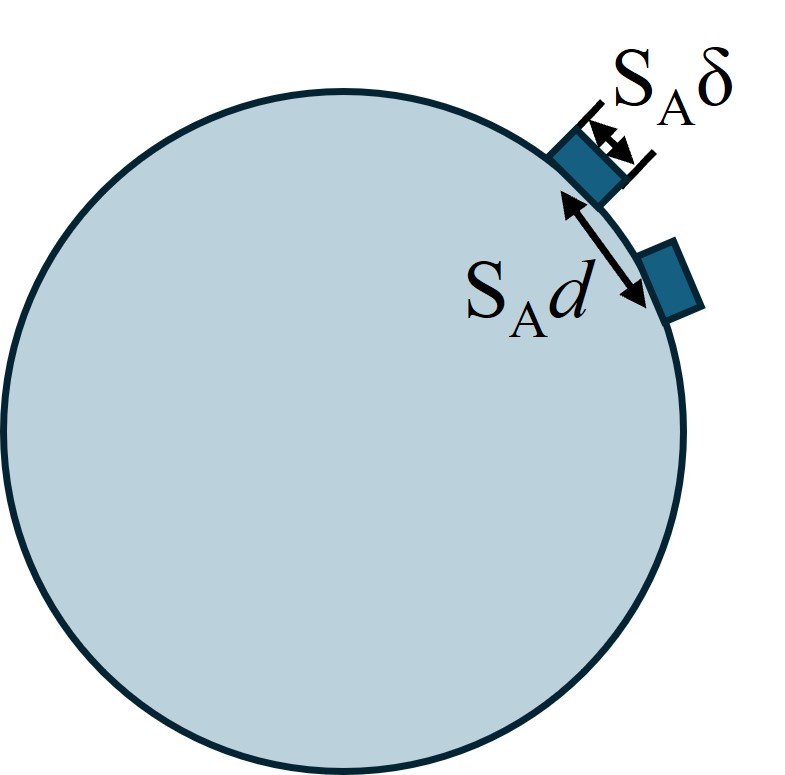}
	}
	
	\subfigure[   ]{
		\includegraphics[height=2.6cm ]{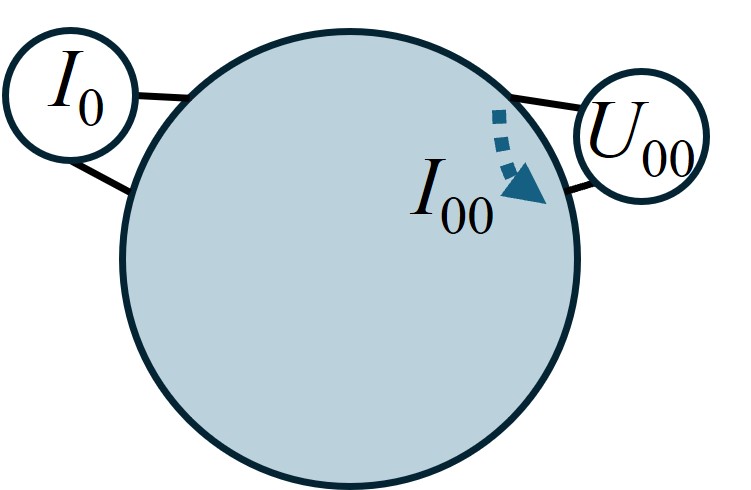}
	}
	\subfigure[  ]{
		\includegraphics[height=2.6cm ]{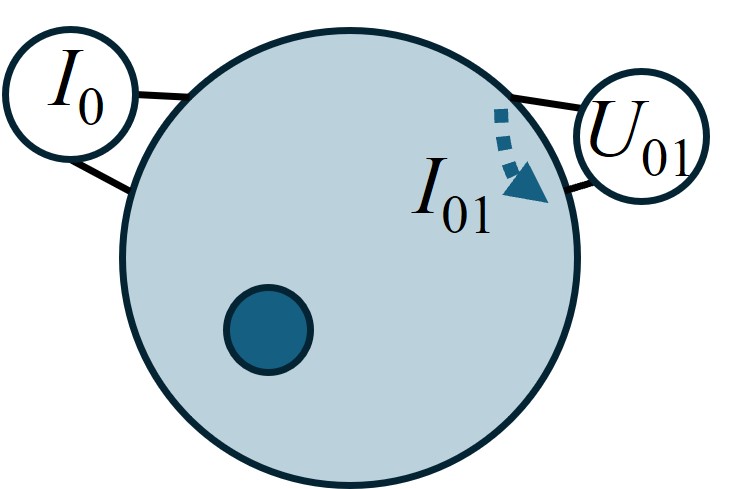}
	}
	
	\subfigure[ ]{
		\includegraphics[height=2.6cm ]{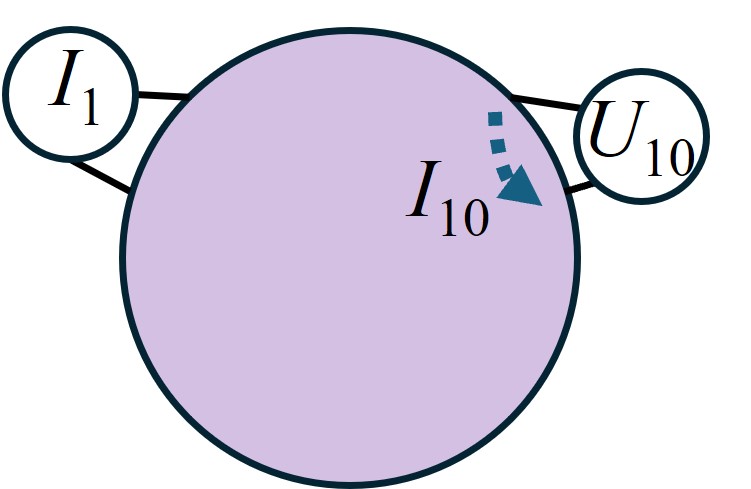}
	}
	\subfigure[   ]{
		\includegraphics[height=2.6cm ]{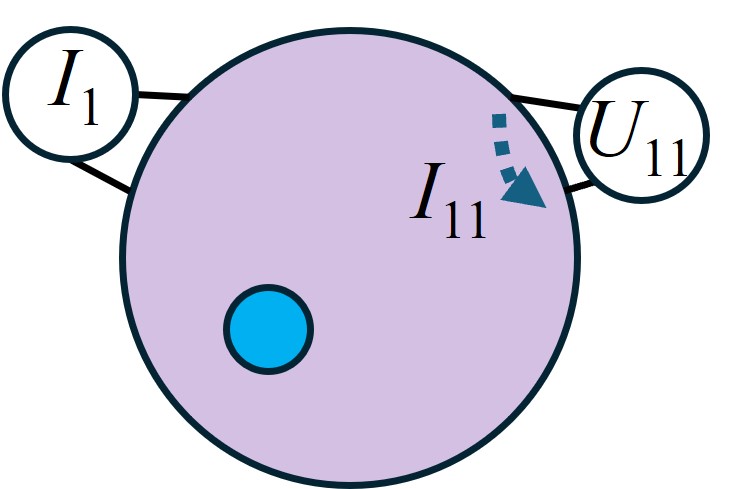}
	}

	\caption{ Variations in size, excitation currents, and background conductivities. (a) Small study area. (b) Study area enlarged by a factor of $S_A$. (c) The case of no inclusions in the simulated data. (d) A case of one inclusion in the simulated data. (e) The case of no inclusions in the real data. (f) A case of one inclusion in the real data. 
	}
	\label{fig.gen}
\end{figure}
 
\section{Model Generalization}
\label{sec.3.gen}
This section presents a generalization method for 2D EIT image reconstruction that enables models trained on simulated data to be applied directly to real data, accommodating variations in size, excitation currents, and background conductivities.

We consider the conventional setting of a circular sensor with the electrodes on the periphery. We first consider the implications of variations in the sensor dimensions (i.e., area) in the context of real-world experiments. 
The equivalent resistance between two points located on the periphery of the disk electrode is expressed as \cite{jeans2009mathematical}:
\begin{equation}
	R\approx \frac{2}{\pi\sigma}\ln \frac{2d}{\delta},\label{eq.resistancee}
\end{equation}
where $d$ represents the length of the chord connecting the two measurement points, and $\delta$ denotes the diameter of the small circular contact area at each point.
Consider a scenario where the geometry of the real dataset is scaled by a factor of $S_A$ compared to the training dataset. 
The resistance value between the electrodes remains constant, as both the chord length and contact width scale proportionally by a factor of $S_A$, as shown in Figs.~\ref{fig.gen} (a)-(b). Consequently, injecting an identical current value through an electrode pair in the real-world setup will yield the same voltage, as observed in the simulation studies. 
In other words, an EIT image reconstruction model trained on a unit-length simulated dataset can be directly applied to a real dataset with different, but proportionally scaled, dimensions.

In the remaining of this subsection, we propose and demonstrate the application of normalization techniques to address scenarios where the real-world experimental setup introduces variations in both the values of the excitation current and the background conductivity level.

To demonstrate the principle, we consider only voltage changes across a pair of neighboring electrodes.
In the case of the simulated dataset, $I_0$ current is injected on the excitation  electrodes.
We assume that the effective current values between the measurement electrodes are $I_{00}$ and $I_{01}$ for the target area without and with inclusions, and the corresponding conductivity distributions are 
$\sigma_{00}$ and $\sigma_{01}$, respectively.
Based on Ohm's law, the values of the currents in the case of the simulation study are:
\begin{equation}
\begin{array}{l}
	I_{00}=U_{00}  \sigma_{00}\\
	I_{01}=U_{01}  \sigma_{01}
\end{array}
\end{equation}
where $U_{00}$ and $U_{01}$ are the boundary voltages in the absence and presence of an inhomogeneity, as shown in Figs.~\ref{fig.gen} (c)-(d).

The trained model $g(\cdot)$ can be used to predict the change of the conductivity via $\delta\widehat{ \bs{\sigma}}_0=g(U_{01}-U_{00})$. For the real world dataset cases, shown in Figs.~\ref{fig.gen} (e)-(f), the current formulae are identical: 
\begin{equation}
	\begin{array}{l}
	I_{10}=U_{10} \sigma_{10}\\
	I_{11}=U_{11} \sigma_{11}
\end{array}\label{eq.real}
\end{equation}
where $U_{10}$ and $\sigma_{10}$, $U_{11}$ and $\sigma_{11}$ are, respectively, the boundary voltages and conductivity distributions, in the absence and presence of an inhomogeneity.

Due to the domain translation, direct deployment of the EIT image reconstruction model trained on the simulated data to the real data will 
result to incorrect conductivity estimates, $\delta\widehat{ \bs{\sigma}}_1\neq g(U_{11}-U_{10})$.
However, the proposed normalization procedure can be used to tackle the model mismatch between simulated and real-world datasets.

In specific, the current scaling factor $S_I$ is determined to be $S_I=I_1/I_0=I_{10}/I_{00}=I_{11}/I_{01}$ due to linearity.
This factor can be used to normalize the values of the excitation current in \eqref{eq.real}: 

\begin{equation}
	\begin{array}{l}
	I_{00}=U_{10} \sigma_{10}/S_I \\
	I_{01}=U_{11} \sigma_{11}/S_I 
\end{array} 
\end{equation}
Next, the additional term $U_{00}/U_{10}\cdot U_{10}/U_{00}$ is introduced to the above equation, which can be simplified to: 

\begin{equation}
	\begin{array}{l}
	I_{00}=U_{10}\frac{U_{00}}{U_{10}}\cdot  \sigma_{00}  \\
	I_{01}=U_{11}\frac{U_{00}}{U_{10}} \cdot \sigma_{11} \frac{U_{10}}{U_{00}} \frac{ 1}{S_I} 
\end{array} \label{eq.real.1}
\end{equation}
We proceed to introduce the voltage scaling factor as  $S_U=U_{10}/U_{00}$.
This leads to \eqref{eq.real.1} being further simplified as: 
\begin{equation}
	\begin{array}{l}
		I_{00}=U_{10}\frac{1}{S_{U}}\cdot  \sigma_{00} \\
		I_{01}=U_{11}\frac{1}{S_{U}} \cdot \sigma_{11} \frac{ S_U }{S_I} 
	\end{array} \label{eq.real.2}
\end{equation}

As the real dataset has been normalized to the case where the conductivity of the background is $\sigma_{00}$ and the excitation current  is $I_{0}$, the model trained on the simulated dataset can be used to predict the normalized conductivity changes as:
\begin{equation}
	\delta\widehat{ \bs{\sigma}}_1=g(\frac{U_{11}-U_{10}}{S_U})\cdot\frac{S_I}{S_U}
\end{equation}
In the experiments performed in this work, the voltage scaling factor is approximated by $S_U\approx  \mathbf{\overline U}_r    / \mathbf{\overline U}_s  $, where $\mathbf{\overline U}_r $
and $\mathbf{\overline U}_s$ are the average values of the reference voltages without the presence of inclusions for the simulated and real data, respectively.

\begin{figure}[!t]
	
		\centering
 
	\subfigure[]{
		\includegraphics[height=4cm]{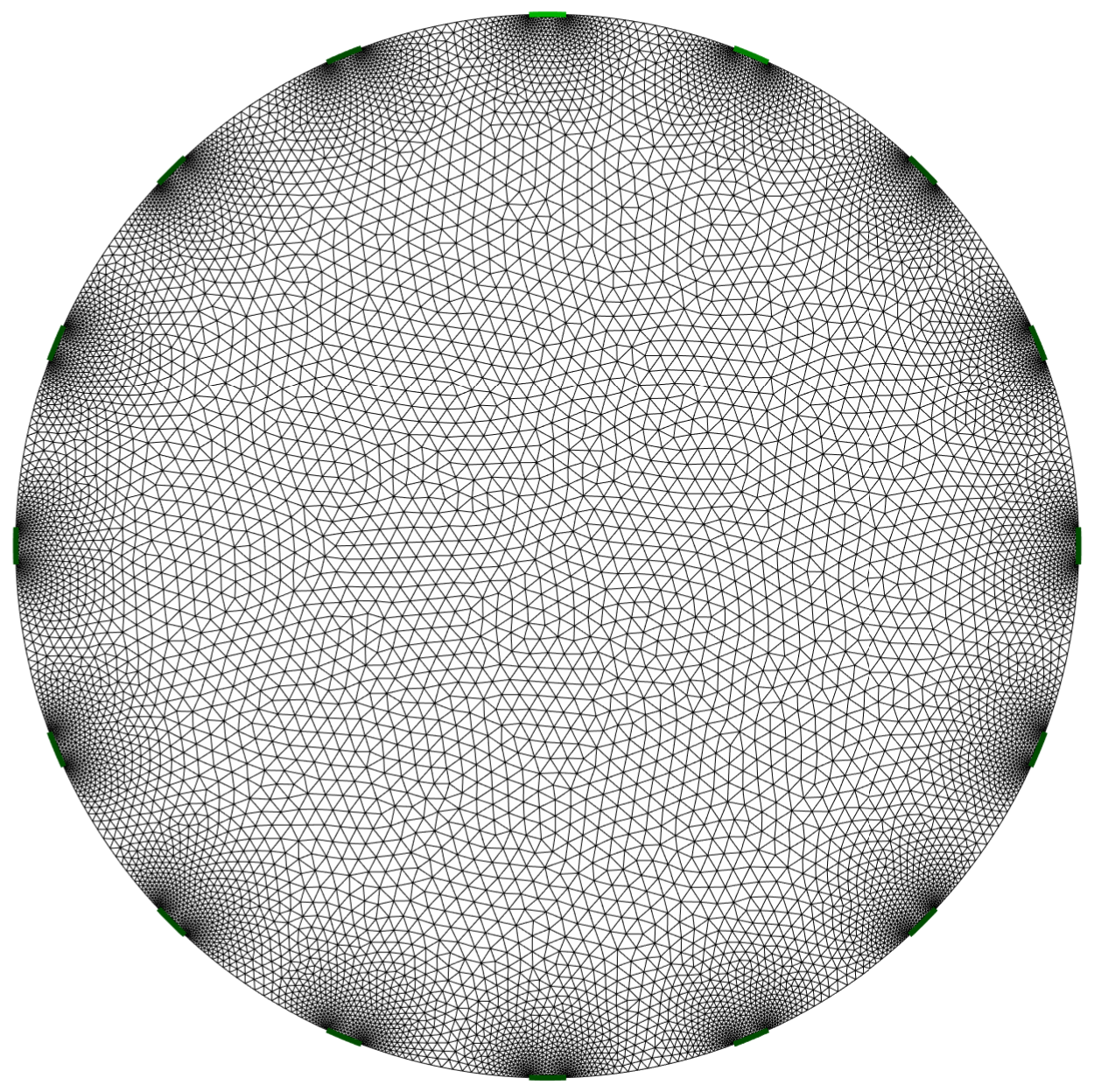}
	} 
	\subfigure[]{
		\includegraphics[height=4cm]{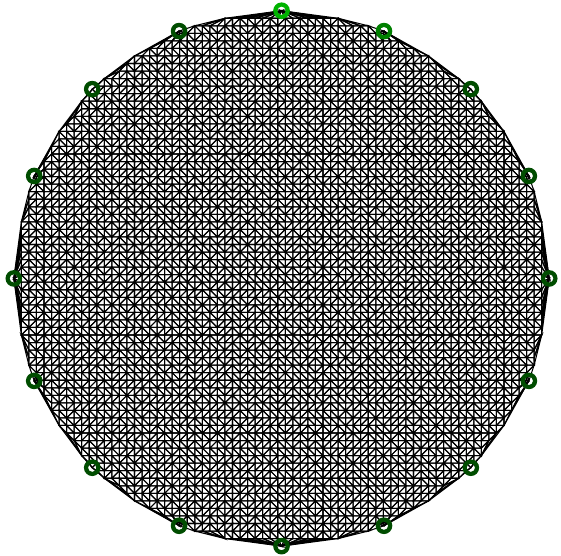}
	} 
	\caption{Forward and inverse FEM mesh configurations. (a) Forward mesh. (b) Pixelized inverse mesh.
	}
	\label{fig.mesh}
\end{figure}

\section{Experiments}
\label{sec.sec4}
We present the experimental results of the proposed CDEIT conducted on one simulated dataset and two publicly available real datasets, namely UEF2017\footnote{https://zenodo.org/records/1203914} and KTC2023\footnote{https://zenodo.org/records/10418802}. For comparative analysis, we employ two conventional algorithms, i.e., single-step Tikhonov regression (TR) \cite{TR} and the iterative Gauss-Newton (GN) \cite{GN} method, and several supervised deep learning-based EIT image reconstruction models. These include Improved LeNet \cite{improvedlenet}, CNN-EIM \cite{cnneim}, SADB-Net \cite{sadbnet}, SA-HFL \cite{sahfl}, Ec-Net \cite{ecnet}, DHU-Net \cite{dhunet}, and CSD \cite{csd}.
Unlike the deep learning models, the first two methods, primarily based on the principles of linear regression, struggle to produce accurate conductivity reconstruction images and necessitate the computation of the Jacobian observation matrix. These methods constitute a distinct category, and their results are included for reference purposes.

\subsection{Data Description}
\begin{figure}[!t]

	\hspace{0.4cm}
	\subfigure[]{
		\includegraphics[height=2.7cm]{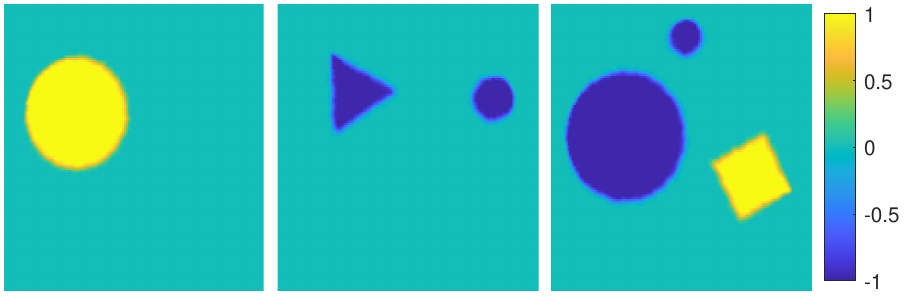}
	} 
	
	\centering
	\subfigure[]{
		\includegraphics[height=2.2cm]{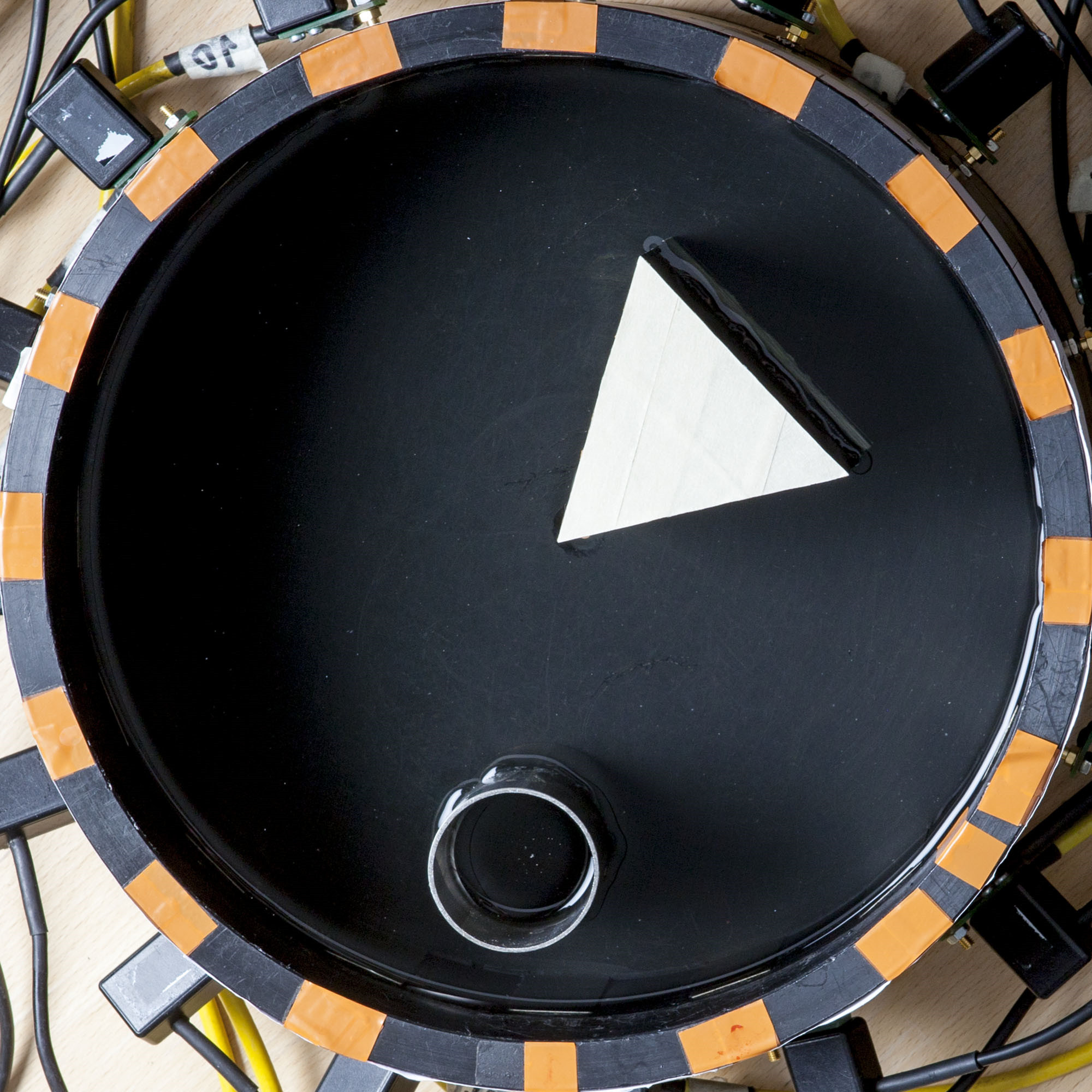}
	} 
	\subfigure[]{
		\includegraphics[height=2.2cm]{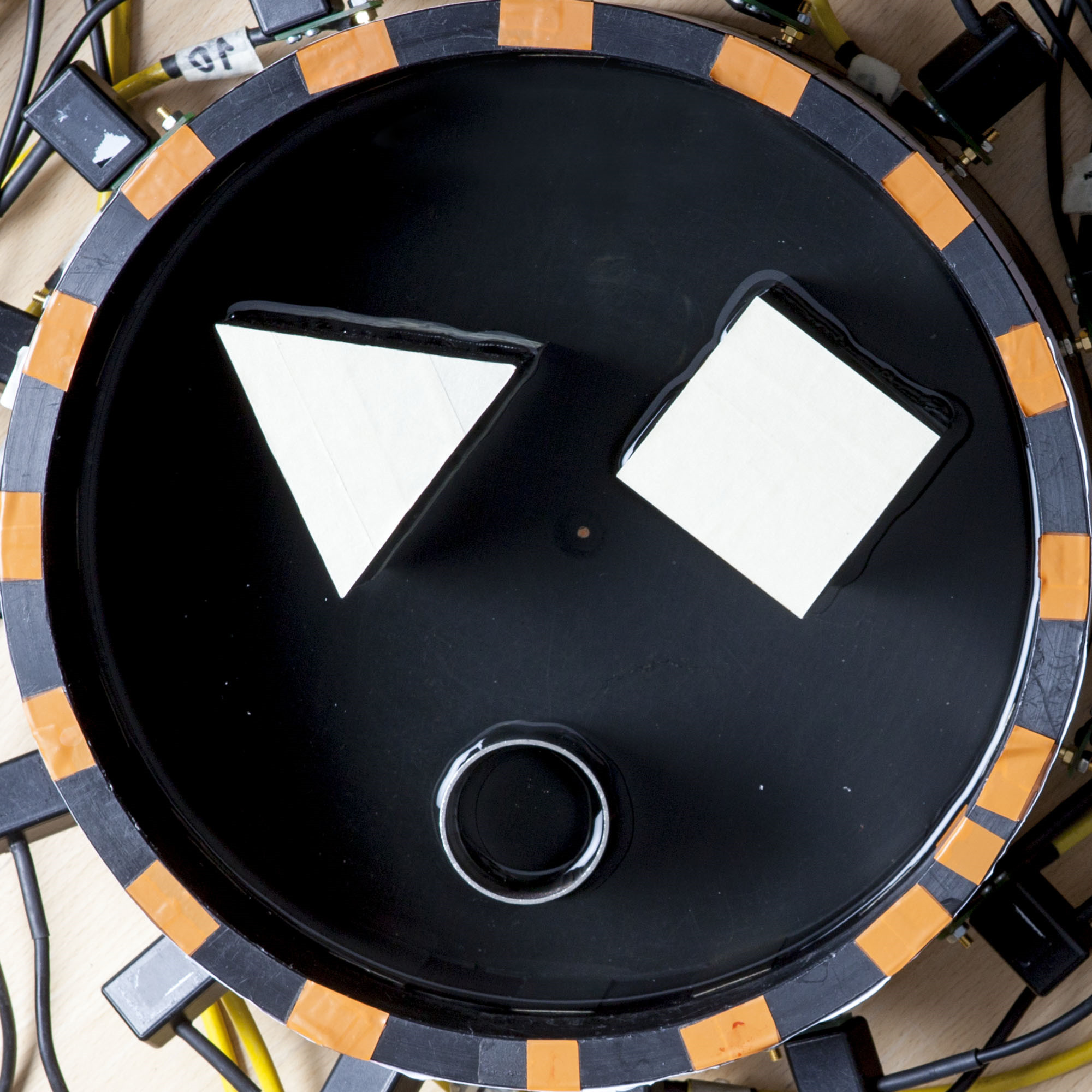}
	} 
	\subfigure[]{
		\includegraphics[height=2.2cm]{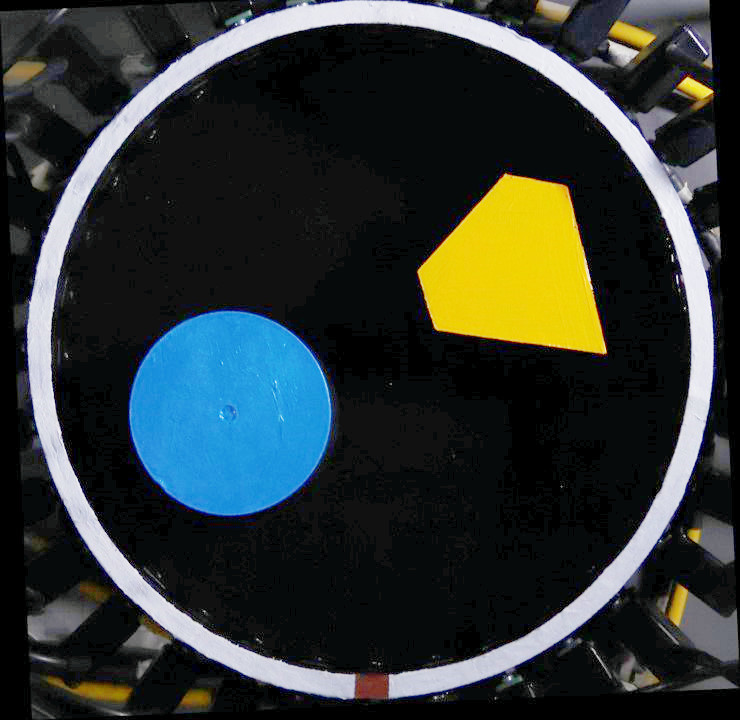}
	} 
	\caption{Example benchmark  images in the experimental datasets. (a) Conductivity changes for three samples from the simulated dataset, with one, two and three objects, respectively (b)-(c) \textit{Samples 4\_1} and \textit{5\_1} from the UEF2017 dataset, respectively, (d) a sample from the KTC2023 dataset.
	}
	\label{fig.benchmarkimages}
\end{figure}

The simulated dataset consists of a number of inhomogeneities of varying shape, i.e., circular, triangular and square. We explored combinations between one and three objects. The background conductivity is set to 1 S/m. The inclusions are randomly designated as either insulating or conductive materials, with conductivities of 0.01 S/m and 2 S/m, respectively. All simulation experiments were performed using the MATLAB's EIDORS \footnote{https://eidors3d.sourceforge.net} open-source toolbox \cite{EIDORS}, which provides a variety of choices, including sensor geometry, current injection and voltage measurement protocols, and material properties.
We followed the popular measurement setup  \cite{Sheffield} of a circular sensor with 16 electrodes on the boundary, adjacent current stimulation of 1 mA, and the adjacent voltage measurement protocol.
Next, the finite element method (FEM) was used to solve the EIT forward problem in \eqref{eq.eit.f1}-\eqref{eq.eit.f4}. This resulted to boundary voltage measurement vectors with 208 elements for each data sample. Fig.~\ref{fig.mesh} illustrates meshes for solving the forward and inverse problems. For the forward problem using the circular homogeneous model, the mesh comprises 10,496 nodes and 20,230 elements. The mesh for the inverse problem solved by TR and GN is set to a regular shape and consists of 3,956 nodes and 7,862 elements. Next, the values of the FEM elements were interpolated to a pixel space of size  128 $\times$ 128   for the training of deep learning-based models and performance evaluation for all methods.
The simulated data is randomly partitioned into training, validation, and test sets with a ratio of 8:1:1. The resulting datasets contain 46,080, 5,760, and 5,760 samples, respectively.
Three benchmark conductivity images are illustrated in Fig.~\ref{fig.benchmarkimages} (a).

The second dataset, UEF2017\cite{uef2017}, was collected in 2017 at the University of Eastern Finland. The experiments were performed using a circular, flat water tank with a radius of 14 cm.
There were sixteen electrodes uniformly distributed around the periphery of the tank with an excitation current of 2 mA. The interior of the tank was filled with  saline solution to a height of 7 cm,
containing various conductive and resistive inclusions. Samples with uneven backgrounds and complex inclusions were removed, and the remaining 22 samples were used for the evaluation.
Fig.~\ref{fig.benchmarkimages} (b) and (c) depict the RGB images of two samples from the dataset. In the absence of ground truth for the changes in the conductivity distributions, these images were utilized to generate the reference conductivity. In this setup, the white inclusions represent insulators with a conductivity change value of -1, while the black inclusions represent conductors with a conductivity change value of +1. This configuration was used for evaluation purposes.

The third dataset is KTC2023 \cite{ktc2023} from the Kuopio Tomography Challenge held in 2023.
The measurement system comprises an electrical impedance tomography device with a circular plastic water tank, and 32 electrodes uniformly distributed along its periphery.
Half of the electrodes were used to inject a current of 2 mA sequentially in the skip-1 manner\cite{skip1}, i.e., injecting current to $1-3, 3-5, \dots , 31-1$ electrodes.
The voltages between all neighboring electrodes were collected.
When the voltage values of two adjacent pairs of electrodes are summed together, they yield voltage data equivalent to using adjacent current excitation and adjacent voltage measurement patterns with 16 electrodes.
This dataset incorporates conductivity change masks, which were utilized for evaluation purposes. The reference conductivity change parameters align with those employed in the UEF2017 dataset.

\subsection{Experimental Setup}

\subsubsection{Hyperparameter Settings}

The Transformer-based Unet is used for the conductivity reconstruction.
The forward diffusion time step $T$ is set to  1000. 
The hyperparameter sequence $\{\beta_1,\beta_2,\dots,\beta_T\}$ is set to a sequence with uniform growth from $ 1\times e^{-4} $ to $ 2\times e^{-2} $. 
The number of training iterations is set to $150k$, the batch size is set to 64, and the Adam optimizer\cite{adam} with learning rate $1\times e^{-5}$ is adopted here.
Once the model training is complete, the image reconstruction is performed using a DDIM sampler with 5 sampling steps.
For the deep learning-based methods, we performed the image reconstruction task on an NVIDIA Tesla V100 GPU with 32GB memory, while the two conventional algorithms were run on an
Intel i7-10870H CPU with 16GB of RAM.

\newcommand\vsp{\vspace{-0.06cm}}

\subsubsection{Performance Metrics}
The peak signal-to-noise ratio (PSNR),   structure similarity index measure (SSIM) \cite{SSIM} and 
correlation coefficient (CC) are adopted
to quantitatively evaluate the reconstruction results of comparison methods.
PSNR is equivalent to the root mean squared error (RMSE). 
In regards to SSIM, this considers luminance, contrast and structure, and helps to assess perceptual quality changes, e.g., the appearance of edges and texture. CC, on the other hand, provides a pixel to pixel intensity value comparison and supports pattern detection through linear shifting of one image over the other. 
All estimates and ground truth are normalized to $[0,1]$ for the evaluation of  criteria.

 \begin{table}[!t]
 	\centering
        \scriptsize
 	\renewcommand\arraystretch{1.2}
 	\caption{Quantitative metrics of the comparison methods on 5760 test samples from the simulated dataset. The best results are shown in bold, while the second best are underlined.}
 	\label{tab.simulated}
 	\begin{tabular}{c|c|c|c }
 		\toprule[1.3pt]
 		& \multicolumn{1}{c|}{PSNR} & \multicolumn{1}{c|}{SSIM} & \multicolumn{1}{c }{CC} \\ \hline
 		TR \cite{TR}            & \, 15.02 $\pm$ 10.27 & 0.608  $\pm$ 0.247 & 0.797 $\pm$ 0.088 \\ \hline
 		GN    \cite{GN}         & 24.97 $\pm$ 3.88 & 0.852 $\pm$ 0.049   & 0.895 $\pm$ 0.073  \\ \hline
 		Improved LeNet \cite{improvedlenet}& 26.99 $\pm$ 3.13 & 0.923 $\pm$ 0.035    & 0.931 $\pm$ 0.048                    \\ \hline
 		CNN-EIM   \cite{cnneim}     & 30.66 $\pm$ 7.88 & 0.967 $\pm$  0.036   & 0.978 $\pm$ 0.025                    \\ \hline
 		SADB-Net \cite{sadbnet}      & 25.42 $\pm$ 3.60 & 0.848 $\pm$ 0.047    & 0.904 $\pm$  0.063                  \\ \hline
 		SA-HFL\cite{sahfl}         & 38.05 $\pm$ 6.19 & 0.990 $\pm$ 0.015    & {0.990} $\pm$  0.015                   \\ \hline
 		Ec-Net \cite{ecnet}        & 35.51 $\pm$ 5.10 & 0.987 $\pm$ 0.014   & 0.987 $\pm$  0.014                    \\ \hline
 		DHU-Net\cite{dhunet}        & \underline{38.76} $\pm$ \underline{5.31} & \underline{0.994} $\pm$ \underline{0.010}   & \underline{0.992} $\pm$ \underline{0.013}                   \\ \hline
 		CSD\cite{csd}            & 28.19 $\pm$ 5.65 & 0.953 $\pm$ 0.044    & 0.922 $\pm$ 0.133                    \\ \hline
 		CDEIT           & \textbf{39.57} $\pm$ \textbf{4.85}  & \textbf{0.998} $\pm$ \textbf{0.004} & \textbf{0.999} $\pm$ \textbf{0.002}         \\ 
   \bottomrule[1.3pt]   
 	\end{tabular}
 \end{table}

\begin{figure*}[!t]
	\centering
	\includegraphics[width=16cm]{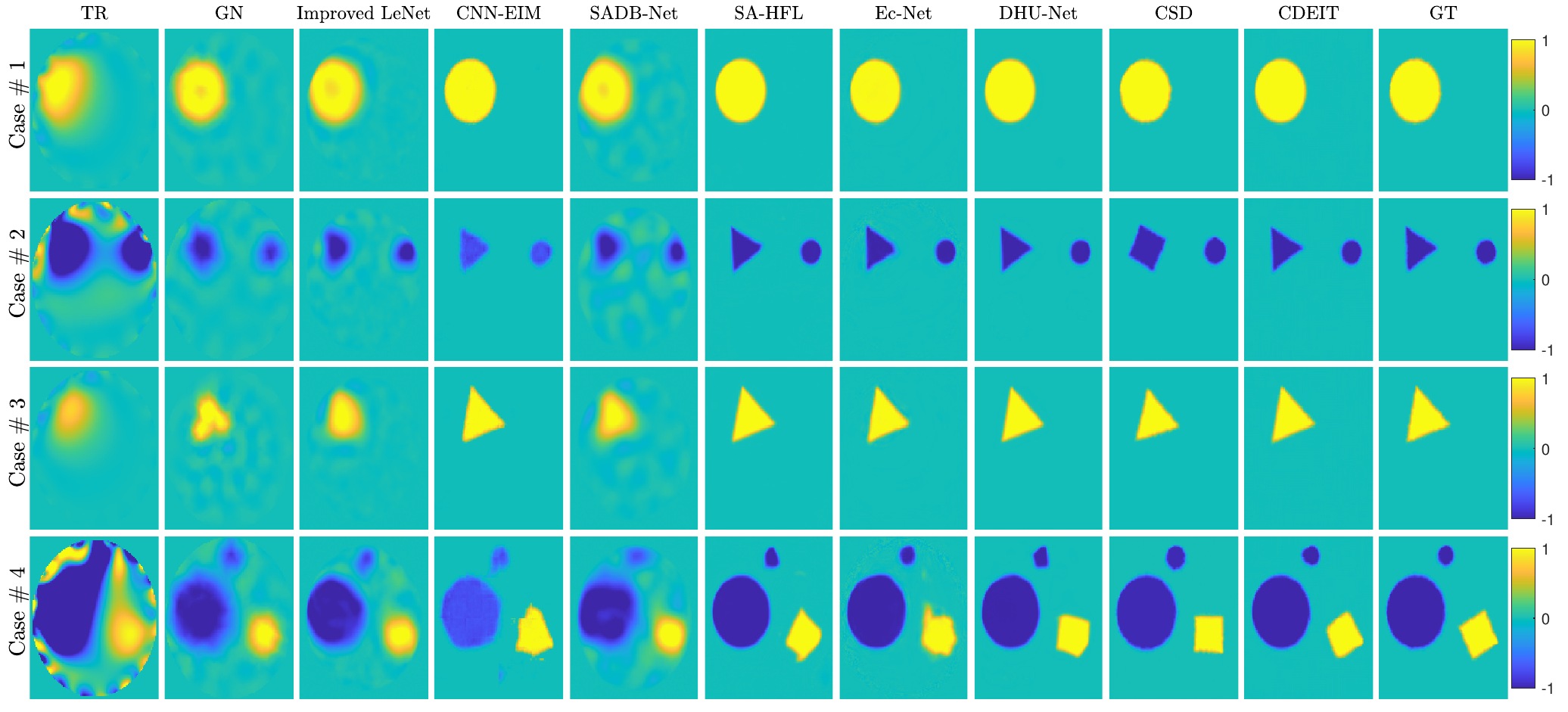}
	\caption{
		Reconstructed conductivity images from simulated test data samples.
	}
	\label{fig.exp.simulated}
\end{figure*}

\subsection{Experiments on the Simulated Dataset}
Table~\ref{tab.simulated} summarizes the  performance metrics of all comparison methods evaluated on the simulated dataset. The results reveal significant variations in reconstruction quality across different approaches.
TR, as a single-step linear reconstruction method, demonstrates the lowest performance overall. In contrast, the GN method, employing an iterative optimization strategy, achieves a substantial improvement, leading to an increase in  PSNR  by 9.95 dB, compared to TR.
Deep learning-based models exhibit competitive performance, compared to  conventional methods. Notably, CNN-EIM demonstrates superior results, yielding a 3.67 dB improvement in PSNR, compared to the Improved LeNet architecture. The dual-branch SA-HFL model, which directly incorporates mask information and integrates more advanced segmentation techniques, outperforms SADB-Net.
Ec-Net and DHU-Net, designed as image refinement models, estimate conductivity images from initial guesses provided by TR. However, their performance is constrained by the limitations of conventional methods, resulting in suboptimal outcomes. 
Moreover, the CSD model functions primarily as an unconditional diffusion model during the training phase, learning the probabilistic distribution of conductivity images. During the prediction stage, it employs a denoising network to process the results generated by the GN method.
This modeling strategy yields a conductivity image with improved clarity, but compromised accuracy. The limitations stem from the direct application of classical reconstruction methods, which, despite their refinement, fail to fully capture the complexities inherent in the EIT inverse problem.

The proposed CDEIT model incorporates numerous reconstruction stages in the reverse denoising process, which are directly conditioned on the voltage signals. We obtain the approximated reconstructed images using the  DDIM  sampler \cite{ddim}.
The performance of the proposed CDEIT model surpasses those of all other reconstruction models evaluated in this study. Notably, it achieves a  {0.81 dB} improvement in PSNR compared to the second-best model, DHU-Net. Moreover, compared to most of the state-of-the-art deep learning models, CDEIT provides more stable image reconstruction, as manifested by the relatively low standard deviation of the performance metrics.
This enhancement underscores the efficacy of our approach in addressing the complex inverse problem inherent in electrical impedance tomography. 
As for the SSIM and CC metrics, CDEIT was also 0.004 and 0.006 higher than the second best method, respectively.
 
 \begin{figure*}[!t]
 	\centering
 	\includegraphics[width=16cm,height=7cm]{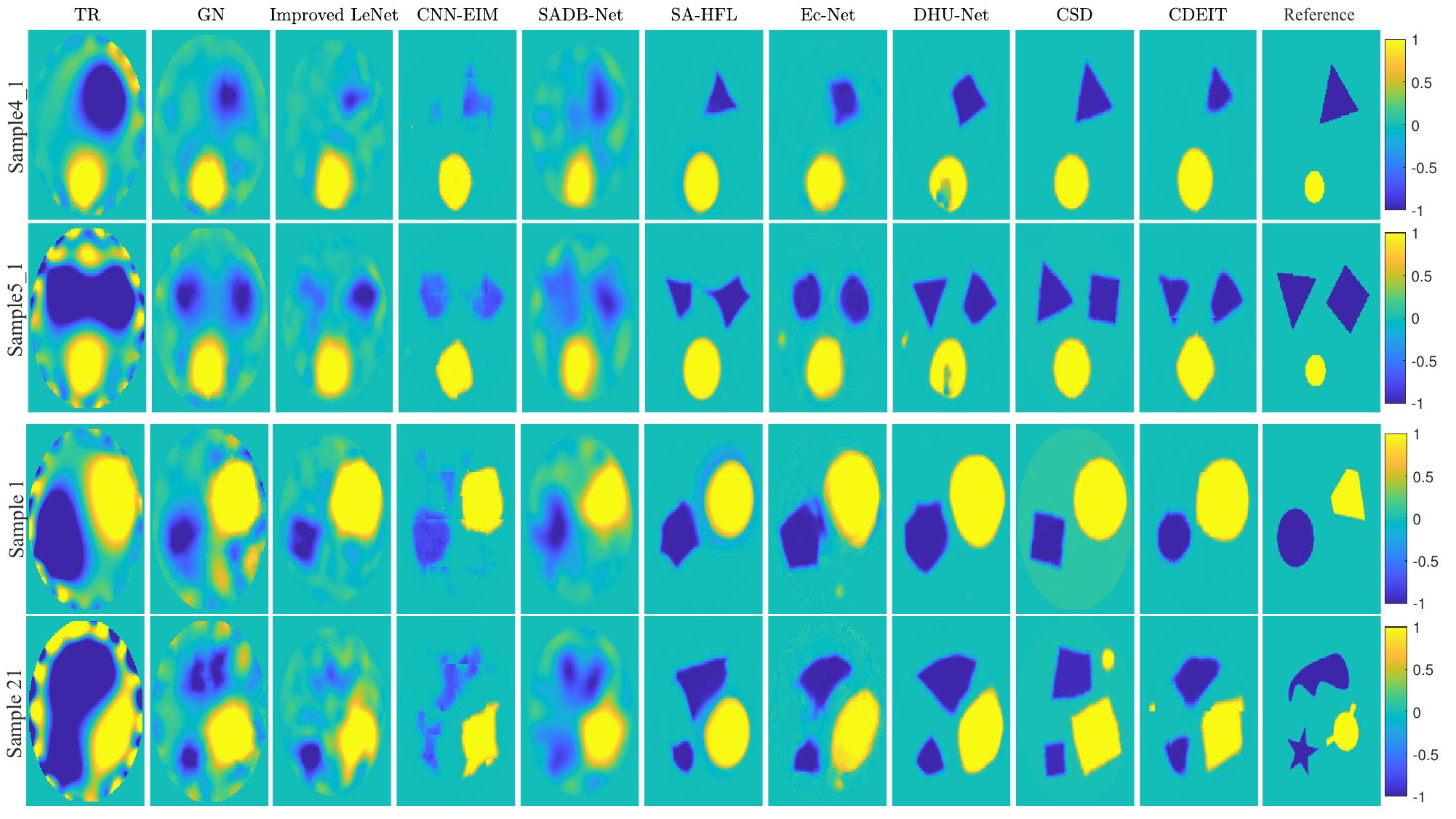}
 	\caption{
 		Reconstructed conductivity images from two real test data samples.
 	}
 	\label{fig.exp.data2017}
 \end{figure*}

Fig.~\ref{fig.exp.simulated} illustrates the reconstructed conductivity images derived from the simulated test dataset, showcasing the comparative performance of various reconstruction methods.
As illustrated in the first column of Fig.~\ref{fig.exp.simulated}, the reconstructed images generated by the TR method fail to accurately delineate the spatial distribution of the inclusions.
The GN method and improved LeNet model demonstrate the capability to localize the positions of the inclusions, however, they fall short in accurately reconstructing their precise morphology, as evidenced in Case \#3 depicted in Fig.~\ref{fig.exp.simulated}.
The CNN-EIM exhibits high accuracy in image reconstruction scenarios characterized by a limited number of inclusions. However, as the number of inclusions increases, thereby elevating the overall complexity of the reconstruction task, the  performance shows a marked decline as shown in Fig.~\ref{fig.exp.simulated}. This degradation in efficacy is  a consequence of the relatively simple architecture of the CNN-EIM, which may lack the capacity to fully represent and resolve the intricate spatial relationships present in more complex inclusion configurations.
Furthermore, SA-HFL provides clearer results for reconstructing the edges of the inclusions, compared to SADBnet. However, if the segmentation network produces incorrect mask predictions, the model will predict inaccurate locations for conductivity changes, as observed in case \#4 in Fig.~\ref{fig.exp.simulated}.
DHU-Net has a larger capacity than Ec-Net, resulting in more accurate conductivity images for both single and multiple inclusions.

For the diffusion models, CSD relies on the initial estimates of the GN for image reconstruction. Although CSD can reconstruct high-definition conductivity images, it sometimes predicts incorrect shapes for the inclusions. For example, in case \#2 in Fig.~\ref{fig.exp.simulated}, the ground truth in the upper left corner is a triangle, but CSD predicts it as a square.
Evidently, the proposed CDEIT produces the best reconstruction results for the desired conductivity images, compared to other methods, when evaluated against the ground truth.

\subsection{Experiments on Real Datasets}
\begin{figure}[!t]
	\centering
	\subfigure[]{
		\includegraphics[height=6cm,width=8cm]{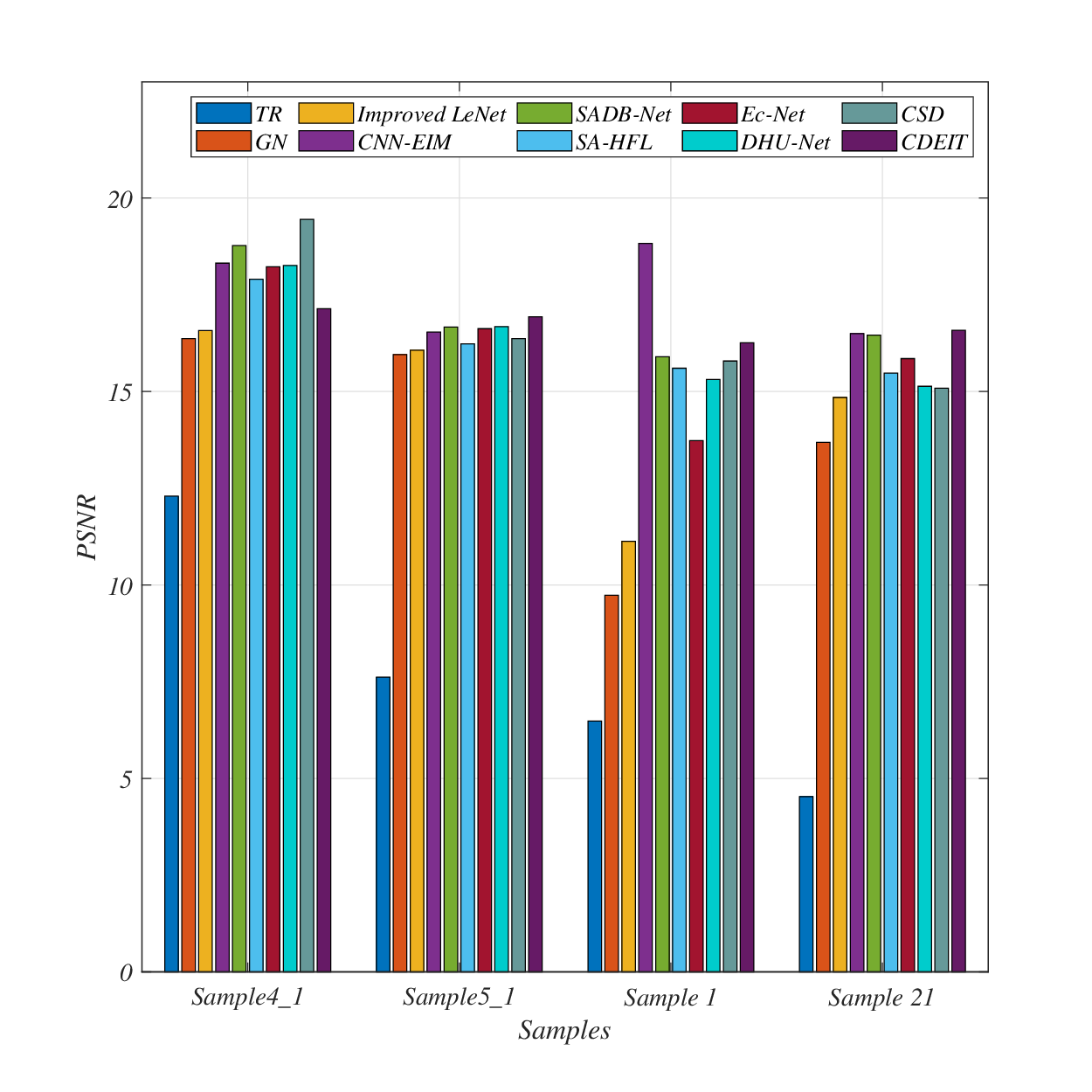}
	} 
	\subfigure[]{
		\includegraphics[height=6cm,width=8cm]{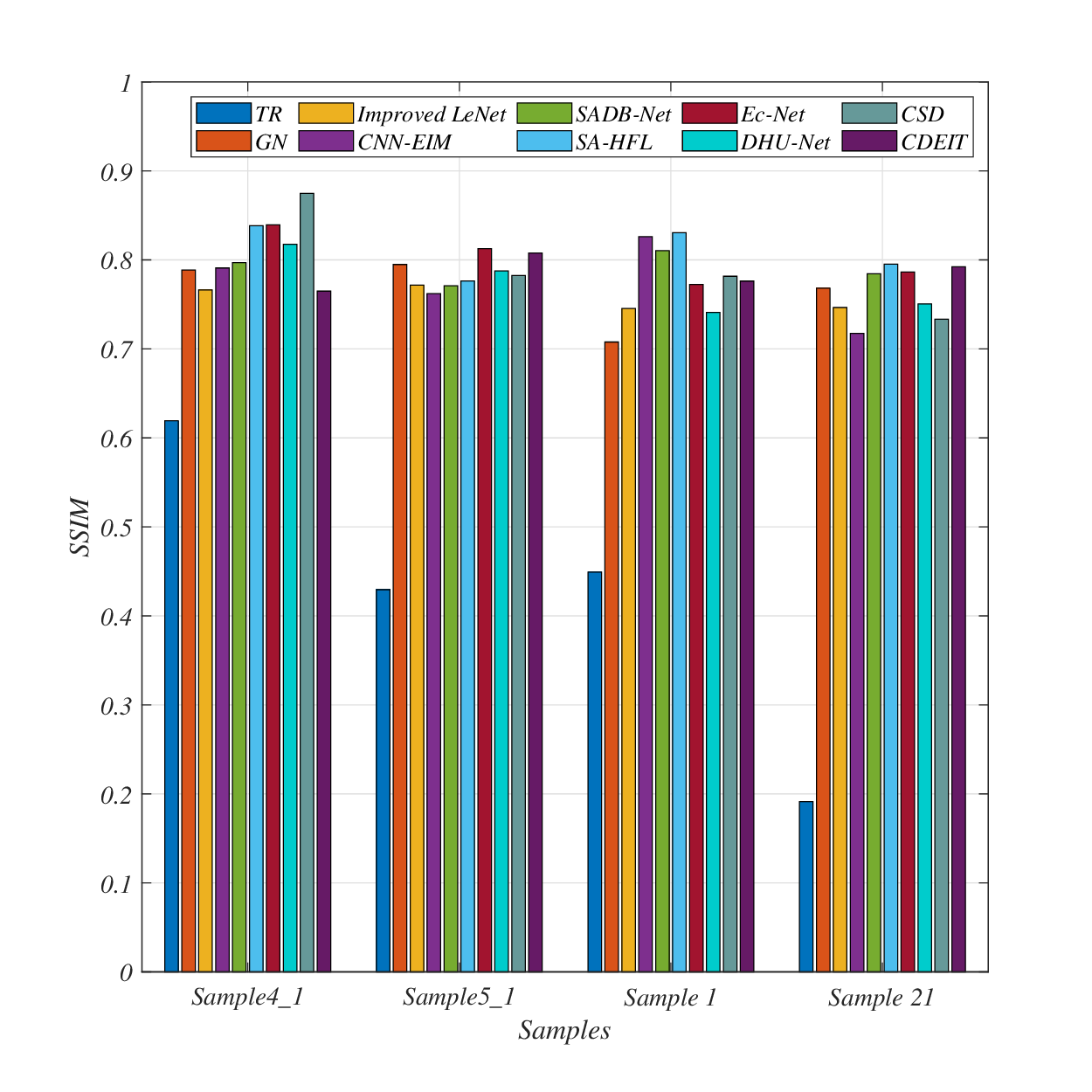}
	} 
	\subfigure[]{
	\includegraphics[height=6cm,width=8cm]{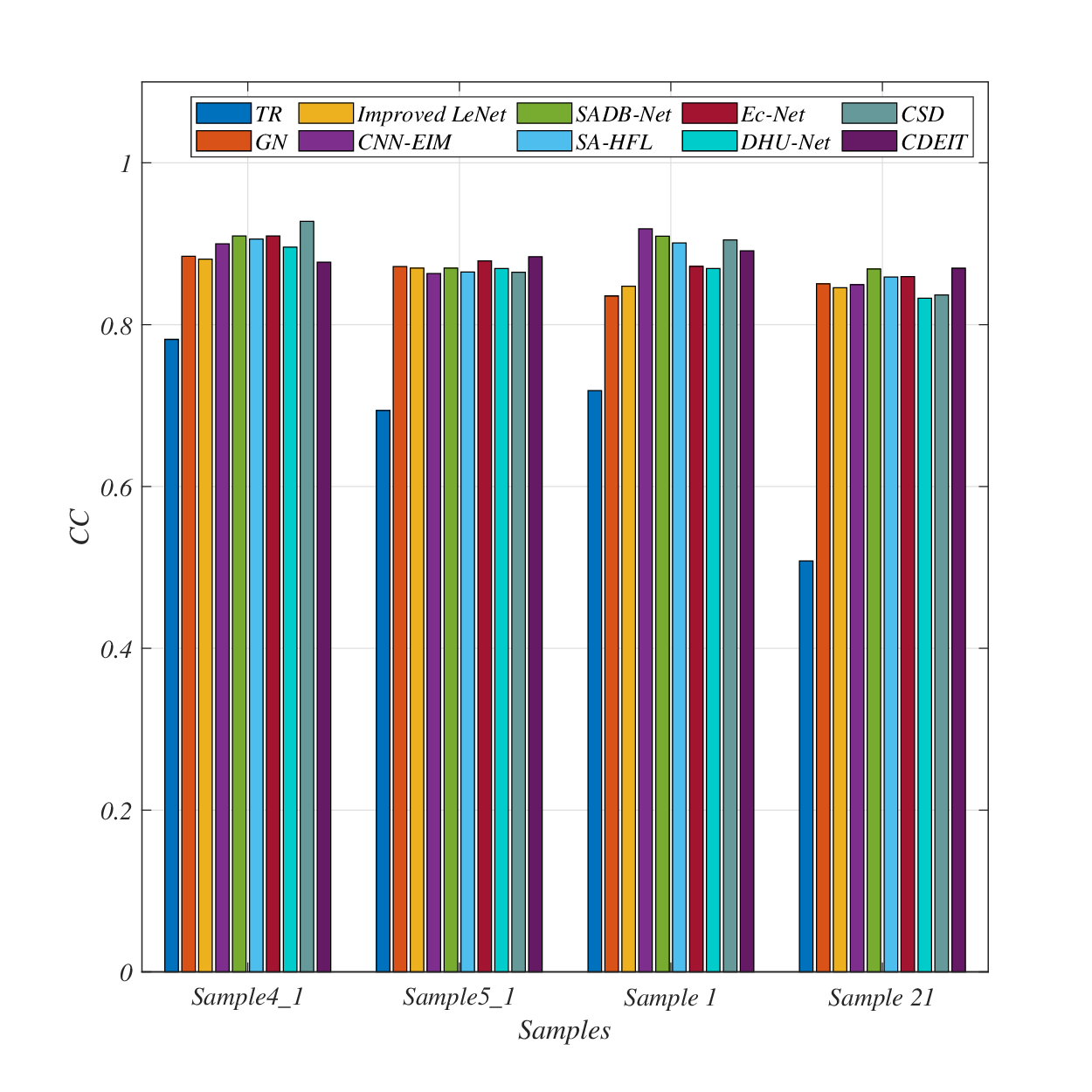}
	} 
	\caption{Reconstruction results for the various methods on 4 samples in real datasets: (a)  PSNR (b) SSIM (c) CC.
	}
	\label{fig.real}
\end{figure}

Furthermore, we conducted experiments on two real EIT datasets with the model trained on the simulated dataset, via the generalization method, described in Sec.~\ref{sec.3.gen},  to further evaluate reconstruction performance. 
Real data, in contrast to the simulated data, have domain shifts caused by measurement noise, electrode movement, domain deformation and temperature effects.
Due to the lack of ground truth conductivity values for the real data, we used manually labeled reference values for the purposes of quantitative assessment.
As can be seen, results tested on these two datasets are significantly worse than the same metrics evaluated on the simulated data in Table \ref{tab.simulated}.
For visual comparison, we present four reconstructed conductivity images with objects characterized by irregular geometry, as seen in Fig.~\ref{fig.exp.data2017}.
The corresponding  metrics  for all image reconstruction methods on these images are shown in Fig~\ref{fig.real}.
The proposed CDEIT is able to obtain competitive performance in terms of the PSNR and CC.
We would like to iterate that in the absence of standardized ground truth for the real data, these quantitative indicators are shown for reference purposes only.

\subsection{Model Discussion}
\label{sec.modelDisscussion}
\subsubsection{Complexity Analysis}
In this subsection, we discuss the computational complexity of the various reconstruction models considered in this research, which relates to the number of parameters and training time/model convergence. 
Table~\ref{tab.complexity} provides an overview of the number of parameters, floating-point operations (FLOPs), when implementing once-forward computing, and training times when using the simulated datasets of the proposed model vs the state-of-the-art DL methods. In regards to model size, which may be thought as being indicative of capacity, the largest models are SADB-Net, and CDEIT. These models could be more suited to handling complex tasks, and may require more resources for training. At the opposite end, Ec-Net and CNN-EIM are lightweight models, which indicates reduced memory requirements. Considering the number of FLOPs, Ec-Net is the most resource-intensive model with 16.35 billion FLOPs, despite having a small parameter size. This is in contrast to CDEIT, which has a large parameter size but relatively few FLOPs (1.55 bilion). Overall, CDEIT strikes a good balance between high model capacity and computational efficiency, offering an effective solution where there are constraints in terms of memory and inference, while model complexity is still important. Moreover, using the DDIM sampler, CDEIT delivers similar performance to multi-stage fusion models. 


\begin{table}[!t]
	\centering
	\renewcommand\arraystretch{1.2}
	\caption{Number of Parameters, FLOPs and training time of comparison models.}
		\label{tab.complexity}
	\begin{tabular}{c|c|c|c}
		\toprule[1.3pt]
 
       &  {Size ($\times   10^{6}$)} &  {FLOPS ($\times   10^{9}$)} &  {Training time (h)} \\ \hline
		 {Improved LeNet}   &  10.13                      & 0.01                         &        0.29       \\ \hline
		CNN-EIM   &  1.73                      &  {0.07}                          &  0.80                    \\ \hline
		SADB-Net   &  176.52                       &  {11.38}                          &  28.98                   \\ \hline
		SA-HFL  &  7.84                       &  {4.35}                          &  25.29                   \\ \hline
		Ec-Net   &  1.00                       &  {16.35}                          &  43.48                    \\ \hline
		DHU-Net   &  2.45                       &  0.89                          &  {23.87}                   \\ \hline
		CSD  &   4.04                      &  4.15                         &  {5.79}                   \\ \hline
		CDEIT   &  107.52                       &  1.55                          &  43.33                   \\  

			\bottomrule[1.3pt] 
	\end{tabular}
\end{table}

\subsubsection{Ablation Study}
We performed an ablation study for the EIT image reconstruction task, exploring different types of  normalization layers and activation functions to determine the optimal settings. 
This experiment was conducted on the simulated dataset to save time by only  using $10k$  training steps. Table~\ref{tab.ablation} presents the PSNR results, when using layer normalization and RMS normalization, with SiLU and ReLU activation functions. It is evident that these configurations achieve very similar reconstruction performance. We adopted the use of SiLU and RMSnorm in the proposed model due to their simplicity and absence of additional hyperparameters.

\begin{table}[]
		\centering
	\renewcommand\arraystretch{1.2}
	\caption{Ablation experiments for different configurations.}
	\label{tab.ablation}
	\begin{tabular}{c|c|c}
		\toprule[1.3pt] 
		 
	 \multirow{2}{*}{ReLU} & LayerNorm      & 17.10   \\ \cline{2-3} 
		                      & RMSNorm & 17.10   \\ \cline{1-3} 
	 \multirow{2}{*}{SiLU} & LayerNorm      & 17.11 \\ \cline{2-3} 
		                       & RMSNorm & 17.12 \\ 	\bottomrule[1.3pt] 
	\end{tabular}
\end{table}

\subsubsection{Robustness Analysis}
Last, we tested the generalization ability of the EIT reconstruction models.
Gaussian white noise was introduced at 40 dB and 30 dB in the simulated test set, and the new datasets are labeled as "Data 40 dB" and "Data 30 dB", respectively. The original dataset is labeled as "Data". 
The PSNR values of the image reconstruction results at different noise levels are shown in Table \ref{tab.robust}.
From the results, it can be seen that when the amount of noise is low, CDEIT achieves high reconstruction accuracy, however as the amount of noise increases, its performance deteriorates compared to the DHU-Net and Ec-Net models, which rely on the classical TR approach.

\begin{table}[!t]
	\centering
	\renewcommand\arraystretch{1.2}
	\caption{PSNR of image reconstruction results at different noise levels on the simulated dataset.}
	\label{tab.robust}
	\begin{tabular}{c|c|c|c}
		\toprule[1.3pt]
		& Data  & Data 40 dB & Data 30dB \\ \hline
		TR             & 15.02 & 14.93      & 14.28     \\ \hline
		GN             & 24.97 & 22.69      & 18.60     \\ \hline
		Improved LeNet & 26.99 & 26.08      & 22.99     \\ \hline
		CNN-EIM        & 30.66 & 28.31      & 23.40     \\ \hline
		SADB-Net       & 25.42 & 24.32      & 19.46     \\ \hline
		SA-HFL         & 38.05 & 33.20      & 25.11     \\ \hline
		Ec-Net         & 35.51 & 32.41      & \underline{28.78}     \\ \hline
		DHU-Net        & \underline{38.76}  & \underline{34.35}     & \textbf{28.94}     \\ \hline
		CSD            & 28.19 & 26.02      & 22.36     \\ \hline
		CDEIT           & \textbf{39.57} & \textbf{35.58}      & 26.70     \\
		\bottomrule[1.3pt] 
	\end{tabular}
\end{table}

\section{Conclusion}
\label{sec.sec5}
In this research, we introduce CDEIT, a novel supervised EIT image reconstruction method based on conditional denoising diffusion. This method leverages spatial details in conductivity images through a learned conditional generative model. A Transformer-based U-net is used to model the denoising process in the reverse diffusion model.
Following training, CDEIT is able to generate EIT images from the conditional inputs of the voltage measurements by performing conditional denoising through a step-by-step reverse process. Moreover, a new methodology is introduced to support generalizion of the application of EIT image reconstruction models trained on simulated data on real-world data with varying background conductivities and excitation currents.
Experiments conducted on one simulated and two publicly available datasets demonstrate the effectiveness and efficiency of the proposed model.

Future research will consider  incorporating pre-trained diffusion models from RGB images to enhance the quality of reconstructed conductivity distributions. Additionally, we plan to explore diffusion models operating in lower-dimensional latent spaces to reconstruct high-resolution conductivity images and further improve model efficiency.


\appendices
\section*{Appendix}
In this section, we present the derivation of the evidence lower bound (ELBO) \eqref{eq.elboKL}.

In the DDPM model, all noise-corrupted images, $\bs{\sigma}_1,\bs{\sigma}_2,\dots,\bs{\sigma}_T$, are seen as latent matrices.
From the framework of variational inference,
the real posterior distribution $p(\bs{\sigma}_{1:T}|\bs{\sigma}_0)$ is approximated by a variational distribution $q(\bs{\sigma}_{1:T} |\bs{\sigma}_{0})$, where $\bs{\sigma}_{1:T}$ represents  $\bs{\sigma}_1,\bs{\sigma}_2,\dots,\bs{\sigma}_T$ for short.
Generally, we minimize the KL divergence between these two distributions as
\begin{align}
	&\text{KL}(q(\bs{\sigma}_{1:T}|\bs{\sigma}_{0})||p(\bs{\sigma}_{1:T}|\bs{\sigma}_0))\nonumber\\
	=&\int q(\bs{\sigma}_{1:T}|\bs{\sigma}_{0}) \log \frac{q(\bs{\sigma}_{1:T}|\bs{\sigma}_{0})}{p(\bs{\sigma}_{1:T}|\bs{\sigma}_0)}d\bs{\sigma}_{1:T}\nonumber\\
	=&\int q(\bs{\sigma}_{1:T}|\bs{\sigma}_{0}) \log \frac{q(\bs{\sigma}_{1:T}|\bs{\sigma}_{0})p(\bs{\sigma}_{0})}{p(\bs{\sigma}_{0:T})}d\bs{\sigma}_{1:T}\nonumber\\
	=&\int q(\bs{\sigma}_{1:T}|\bs{\sigma}_{0}) \log \frac{q(\bs{\sigma}_{1:T}|\bs{\sigma}_{0})}{p(\bs{\sigma}_{0:T})}d\bs{\sigma}_{1:T}+\log p(\bs{\sigma}_{0})\nonumber\\
	=& \log p(\bs{\sigma}_{0})-\mathbb{E}_q \left [\log \frac{p(\bs{\sigma}_{0:T})}{q(\bs{\sigma}_{1:T}|\bs{\sigma}_{0})}\right ]
\end{align}
Due to the nonnegativity of KL divergence,
we get the ELBO of the log-likelihood as
\begin{align}
\mathbb{E}_q \left [\log \frac{p(\bs{\sigma}_{0:T})}{q(\bs{\sigma}_{1:T}|\bs{\sigma}_{0})}\right ] \leq \log p(\bs{\sigma}_{0}).
\end{align}
Therefore, we can maximize the ELBO to achieve the maximum likelihood estimate.
Note that the joint distribution of all variables in the forward diffusion and reverse denoising  are
\begin{align}
	q(\bs{\sigma}_1,\bs{\sigma}_2,\cdots,\bs{\sigma}_T|\bs{\sigma}_0)&=\prod^{T}_{t=1} q(\bs{\sigma}_t|\bs{\sigma}_{t-1}),\\
	 p(\bs{\sigma}_0,\bs{\sigma}_1,\bs{\sigma}_2,\cdots,\bs{\sigma}_T)&=p(\bs{\sigma}_{T})\prod^{T}_{t=1} p_\theta(\bs{\sigma}_{t-1}|\bs{\sigma}_{t}).
\end{align}
Then, the ELBO can be further simplified as 
\begin{align}
	&\mathbb{E}_q \left [\log \frac{p(\bs{\sigma}_{0:T})}{q(\bs{\sigma}_{1:T}|\bs{\sigma}_{0})}\right ]\nonumber\\
	=& \mathbb{E}_q \left [   \log p(\bs{\sigma}_{T})+\sum\limits _{t=1}^T\log   \frac{p_\theta(\bs{\sigma}_{t-1}|\bs{\sigma}_{t})}{q(\bs{\sigma}_{t}|\bs{\sigma}_{t-1})}\right ]\nonumber\\
	=&\mathbb{E}_q \left [   \log p(\bs{\sigma}_{T})+\sum\limits _{t=1}^T\log   \frac{p_\theta(\bs{\sigma}_{t-1}|\bs{\sigma}_{t})}{q(\bs{\sigma}_{t-1}|\bs{\sigma}_{t},\bs{\sigma}_{0})}\cdot\frac{q(\bs{\sigma}_{t-1}| \bs{\sigma}_{0})}{q( \bs{\sigma}_{t}|\bs{\sigma}_{0})}\right ]\nonumber\\
	=& \mathbb{E}_q \left [   \log\frac{p(\bs{\sigma}_{T})}{q( \bs{\sigma}_{T}|\bs{\sigma}_{0})} +\sum\limits _{t=1}^T\log   \frac{p_\theta(\bs{\sigma}_{t-1}|\bs{\sigma}_{t})}{q(\bs{\sigma}_{t-1}|\bs{\sigma}_{t},\bs{\sigma}_{0})}\right ]\nonumber\\
	=& -\text{KL}\left(q(\bs{\sigma}_{T}|\bs{\sigma}_{0})||p(\bs{\sigma}_{T} )\right) \nonumber\\
	&-\sum_{t=1}^T \text{KL}\left(q(\bs{\sigma}_{t-1}|\bs{\sigma}_{t},\bs{\sigma}_{0})||p_\theta(\bs{\sigma}_{t-1}|\bs{\sigma}_{t})\right).
\end{align}
Note that the first term represents the 
KL divergence between the distribution of $\bs{\sigma}_T$, which is the output of the forward diffusion process, and the prior distribution $p(\bs{\sigma}_{T})$. Due to the Gaussian diffusion kernel used in the forward process, the first term is close to 0.
Therefore, we can ignore the first term, making the second term our desired result \eqref{eq.elboKL}.

 
%

\bibliographystyle{bib/IEEEtran.bst}
\bibliography{bib/strings.bib}


	

	

\vfill

\end{document}